\documentclass[11pt,a4paper]{article}

\usepackage[final]{acl}

\usepackage{times}
\usepackage{latexsym}
\usepackage{booktabs}
\usepackage{hyperref}
\usepackage[most]{tcolorbox}
\usepackage{enumitem}
\usepackage{xcolor}
\usepackage{tabularx}
\usepackage{array}
\usepackage{makecell}

\usepackage[T1]{fontenc}

\usepackage[utf8]{inputenc}

\usepackage{microtype}

\usepackage{inconsolata}

\usepackage{graphicx}
\usepackage{float}
\usepackage{capt-of}
\usepackage{threeparttable}
\usepackage{placeins}
\usepackage{caption}
\usepackage{stfloats}

\title{Self- and Other-Labels Induce Bidirectional Bias in LLM Judges}

\author{
  Songeun Chae \quad Min Kim \quad Donghoon Jung \quad Seojin Choi \quad Seohyon Jung\thanks{Corresponding author.} \\
  School of Digital Humanities and Computational Social Sciences, KAIST, South Korea \\
  \texttt{\{songeun, mk, donghoon.jung, seojin64, seohyon.jung\}@kaist.ac.kr}
}

\begin{document}
\maketitle
\begin{abstract}
As LLM-as-a-judge becomes increasingly widespread, self-preference—the tendency of a judge to favor its own outputs—raises growing concerns about evaluation reliability.
However, this bias has been studied predominantly on generated text, where stylistic features and response quality are inevitably conflated.
As a result, existing measurements cannot separate genuine self-preference from these confounds.
We address this limitation by changing the object of evaluation: instead of judging generated text, ten LLMs assess sets of narrative constraints selected from a shared pool, which carry no stylistic fingerprint yet retain a recoverable model-specific signature.
Two experiments on this task yield complementary findings.
Under blind evaluation, self-preference disappears, with a small effect remaining in the opposite direction once selection quality and judge severity are controlled.
Under matched quality, however, self- and other-labels alone—without naming any model—shift scores bidirectionally.
LLM judges inflate scores for self-labeled selections and deflate those for other-labeled ones regardless of the selection’s actual source.
We make two contributions: 1) authorship attribution is a distinct driver of evaluation bias, and 2) ground-truth-free tasks can serve as controlled instruments for studying LLM judge behavior.
\end{abstract}

\section{Introduction}
LLM-as-a-judge has emerged as a widely adopted paradigm for scalable evaluation, using large language models (LLMs) as automatic assessors across diverse tasks \citep{gu2026survey}.
Despite their extensive use, LLM judges have been shown to exhibit systematic biases \citep{shi2025judging, ye2025justice}.
A particularly concerning bias is self-preference \citep{zheng2023judging, xu2024pride}, where an LLM judge rates its own outputs more favorably than those of other models.
Prior work has examined potential sources of this bias, including low perplexity \citep{wataoka2024self} and self-recognition \citep{panickssery2024llm}.
\citet{saraf2025quantifying} investigated the effect of authorship labels by manipulating them across three commercial LLMs, with the \textit{Claude} label elevating scores across judges.

\begin{figure}
    \centering
    \includegraphics[width=1\linewidth]{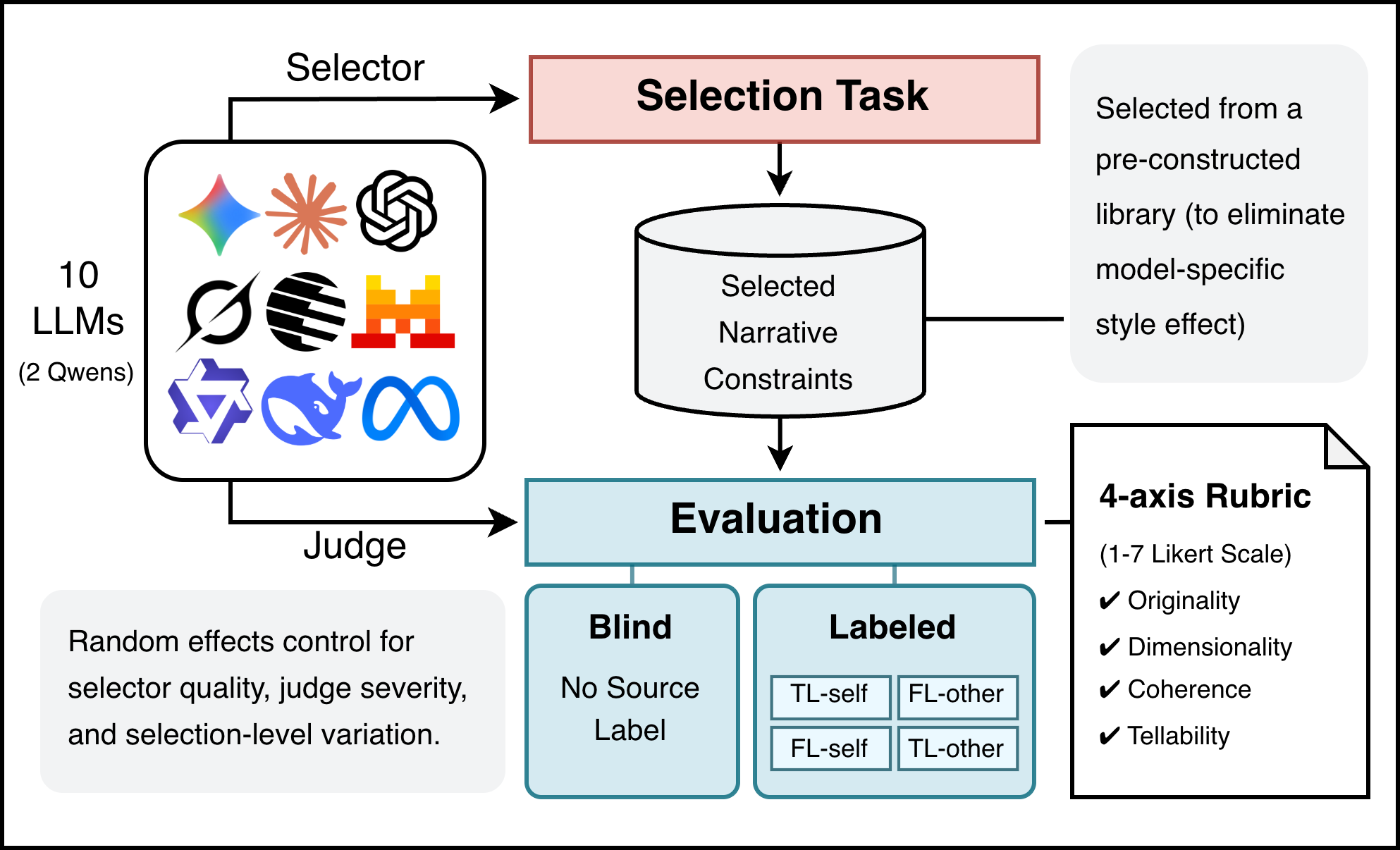}
    \caption{Experimental design. Ten LLMs participate in dual roles: as \textbf{selectors} who construct evaluation targets via a constraint selection task, and as \textbf{judges} who evaluate the resulting selections on a 4-axis rubric. Each selection is judged under two conditions: Blind (no source label) and Labeled, where source attribution is experimentally manipulated in a $2\times2$ design crossing label veracity (TL = true, FL = false) with claimed source (self vs. other).}
    \label{fig:exp_design}
\end{figure}

Meanwhile, recent studies have identified methodological confounds in measuring self-preference.
\citet{chen2025llm, chen2025beyond} argue that existing measurements conflate response quality with genuine self-preference, and \citet{roytburg2026llm} attribute almost 90\% of the observed self-preference to evaluator uncertainty---a judge's limited ability to reliably assess response quality.
These findings raise two questions: First, does self-preference persist once confounds are controlled?
Second, can self/other labels alone shift evaluations under the same controls?

To address these questions, we introduce a novel experimental design built around an open-ended creative task.
Instead of model-generated text, ten LLM judges assess combinations of narrative constraints selected from a pre-constructed pool of 200, under both blind and labeled conditions.
Our results on this task show that although most LLMs assign higher mean scores to their own selections, this apparent self-preference largely disappears once selection quality and judge severity are controlled.
Moreover, authorship labels alone shift evaluations bidirectionally---judges inflate scores under self-labels while simultaneously deflating them under other-labels---even when the evaluation target is identical.

%Our contributions are threefold.
%First, we provide direct evidence that self/other label alone can induce bidirectional evaluation bias in LLM judges.
%Second, we introduce a novel experimental design that structurally eliminates surface confounders, revealing that prior self-preference measurements may have substantially overestimated the effect.
%Third, by leveraging creative tasks as a testbed for assessing LLM judges' biases, we demonstrate that open-ended domains without ground truth can serve as valuable, controlled instruments for studying LLM judge behavior.

We make two contributions.
First, we provide direct evidence that self/other labels alone can induce bidirectional evaluation bias. 
Second, by leveraging creative tasks as a testbed for assessing LLM biases, we show that ground-truth-free settings can serve as controlled instruments for studying LLM judge behavior.
Additionally, our experimental design lends further support to the view that apparent self-preference in LLM judges may be largely a measurement artifact \citep{chen2025beyond, roytburg2026llm}.

\section{Related Work}
\subsection{LLM-as-a-Judge and Evaluation Bias}
LLMs are increasingly deployed as evaluators across a broad range of assessment tasks, such as pairwise comparison and direct scoring \citep{saunders2022self, liu2023g, bai2023benchmarking, li2025generation}.
However, prior work has identified systematic biases that undermine the reliability of LLM-as-a-judge.
Position bias---the tendency of judges to favor a response based on its order rather than its quality---is one of the most extensively studied \citep{zheng2023judging, wang2024large, shi2025judging}.
\citet{koo2024benchmarking} examined six cognitive biases in LLM evaluators, including attentional bias and the bandwagon effect, while \citet{ye2025justice} documented a broader spectrum, showing that judges also prefer responses based on verbosity and sentiment.
Among these biases, \emph{self-preference} \citep{panickssery2024llm, liu2024llms, pombal2026self}---also termed self-enhancement bias \citep{zheng2023judging}---refers to the tendency of an LLM judge to favor its own outputs over other models' outputs.
This bias is particularly concerning because it can directly compromise the validity of LLM-based evaluation, motivating closer examination.

\subsection{Proposed Mechanisms and Measurement Confounds}
While self-preference has been demonstrated across diverse evaluation domains \citep{mahbub2026mitigating, pombal2026self}, its underlying sources remain contested.
\citet{panickssery2024llm} find a positive correlation between self-recognition ability and self-preference strength, whereas \citet{wataoka2024self} attribute self-preference to a familiarity effect, with judges favoring lower-perplexity outputs.
Recent work raises a more fundamental question: whether this effect reflects a genuine bias.
\citet{chen2025beyond} show that existing measures conflate self-preference with response quality; \citet{chen2025llm} argue that, while stronger models exhibit greater self-preference, much of this preference reflects their objectively superior output quality.
Notably, \citet{roytburg2026llm} attribute roughly 90\% of measured self-preference to evaluator uncertainty, demonstrating that judges' difficulty in assessing harder queries inflates the apparent self-preference rate.
Collectively, these findings suggest that most reported self-preference may stem from measurement confounds, though the mechanism behind any residual is still contested.

\subsection{Authorship Labels and Self-Preference}
A distinct line of work examines whether authorship labels can shift LLM-based evaluations.
\citet{sun2026label} reveal that LLM judges attend more closely to the authorship label than to content itself, using it as a shortcut for evaluation.
\citet{marioriyad2025silent} identify a preference hierarchy among authorship labels (Expert $>$ Human $>$ LLM $>$ Unknown).
Most relevant to our work, \citet{saraf2025quantifying} test label-induced bias in three commercial LLM judges under four conditions---no attribution, true attribution, and two false attributions.
The \textit{Claude} label elevates scores while the \textit{Gemini} label lowers them, producing apparent self-preference in Claude and self-deprecation in Gemini.
However, since the evaluated content consisted of LLM-generated blog posts, these effects remain confounded with stylistic fingerprints, perplexity, and output quality, leaving self-recognition as a potential driver of self-preference.
More critically, because the labels were real model names, it is unclear whether the observed shifts reflect self-preference or model reputation effects.

\section{Methodology}
To address the limitations of prior work, we adopt a narrative selection task to construct evaluation targets.
We then prompt LLM judges to assess these selections on a 4-dimension rubric across two settings: blind and labeled evaluation.
Ten LLMs serve as both selectors and judges: five commercial models (Claude Opus 4.7, Gemini 3.1 Pro, GPT-5.5, Grok 4.3, and Qwen3.6-Plus) and five open-weight models (DeepSeek-V4-Pro, Kimi K2.6, Llama 4 Maverick, Mistral Large 3, and Qwen3.6-35B-A3B; see \autoref{app:models}).

\subsection{Narrative Constraint Selection Task}
\label{sec:selection_task}
\begin{figure}
    \centering
    \includegraphics[width=0.95\linewidth]{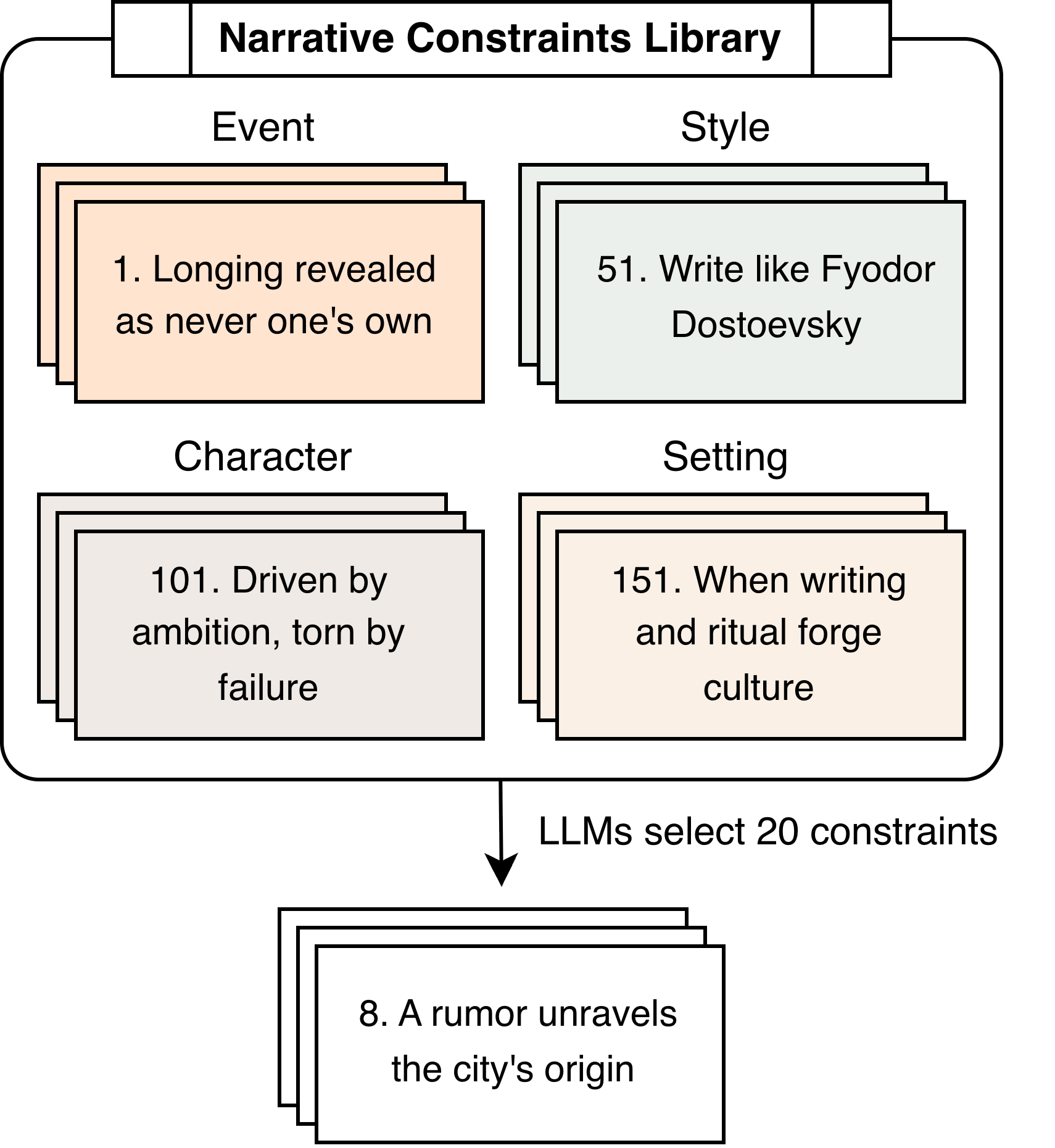}
    \caption{Narrative constraint selection task, adapted from \citet{jung2026style}. The pool comprises 200 single-sentence constraints from that work, organized into four categories. In each run, an LLM freely selects any 20 constraints from the full pool with no category-level quota. All ten models each perform 30 runs, yielding 300 constraint sets in total. No narrative text is generated from the selected constraints; the selections themselves are the evaluation targets.}
    \label{fig:selection_task}
\end{figure}

Instead of relying on text-based evaluation and post-hoc style control, we eliminate model-specific stylistic features by changing the type of evaluation target: LLM judges assess structured narrative selections, with no narrative text generated at any stage.
To produce these selections, we prompt each LLM to perform the selection task 30 times, providing sufficient inter-run coverage per model.
In each run, the LLM selects the 20 items it considers most useful for constructing a single story from a curated set of 200 narrative constraints \citep{jung2026style}---each a pre-written, single-sentence description of an Event, Style, Character, or Setting (see Figure~\ref{fig:selection_task}).
This design offers two advantages.
First, word choice, sentence structure, and phrasing are held constant, preventing model-specific stylistic features from influencing the evaluations.
Second, creative tasks provide a sensitive testbed for detecting LLM judgment biases, as such biases tend to be more salient in subjective, open-ended domains \citep{marioriyad2025silent, fein2026litbench}.

\begin{figure*}[t]
    \centering
    \includegraphics[width=\textwidth]{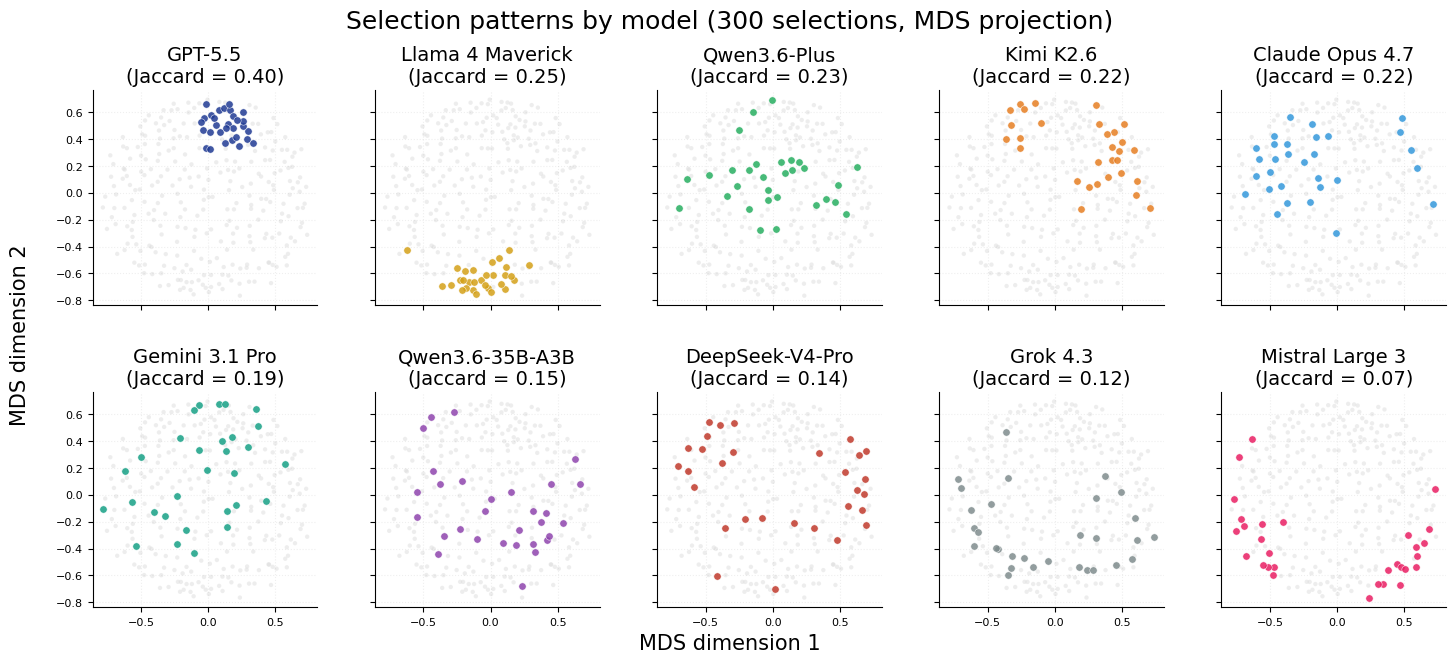}
    \caption{Selection patterns by model (300 selections), projected via Multidimensional Scaling \citep{borg2005modern}. Each subplot highlights one model's selections against the full set; models are ordered by within-model Jaccard similarity (descending).}
    \label{fig:mds_faceted}
\end{figure*}

Analysis of selections reveals that each model draws on a characteristic subset. 
Within-model Jaccard similarity \citep{broder1997resemblance} exceeds chance for all ten models, indicating that each LLM exhibits a consistent selection profile (see \autoref{fig:mds_faceted}; \autoref{app:intra_jaccard} for numerical values).
Moreover, a leave-one-out k-NN classifier ($k=5$) identifies the source model with 50.7\% accuracy---more than five times the 10\% chance baseline (robust across $k \in \{1, 3, 5, 7\}$: 46.7--50.7\%) (see \autoref{sec:appendix_knn_perm}).
Together, these analyses indicate that LLMs maintain distinct selection profiles when model-specific stylistic cues are eliminated by design, providing the variation necessary for meaningful self-versus-other comparisons.

\subsection{Evaluation Procedure}
\subsubsection{Rubric}
Each selection set is assessed on a 7-point Likert scale across four dimensions: \textit{Originality}, \textit{Dimensionality}, \textit{Coherence}, and \textit{Tellability}.
%These dimensions are grounded in narrative theory and chosen to capture properties of narrative selections rather than surface text (see \autoref{app:rubric} for full rubric with anchor descriptions and theoretical rationale).
The rubric provides a structured interpretive frame for eliciting judgments consistently across models and experimental conditions.
\textit{Originality} and \textit{Coherence} are adapted from the Torrance Test of Creative Writing \citep{chakrabarty2024art}, while \textit{Dimensionality} and \textit{Tellability} draw on narrative theory \citep{barthes1977introduction, labov1972language}. 
By directing each judge to assess specific, theoretically grounded properties rather than self-selecting which aspects to attend to, this framework reduces evaluative arbitrariness and renders judge behavior decomposable across conceptually distinct dimensions (see \autoref{app:rubric} for full rubric with anchor descriptions and theoretical rationale).

\subsubsection{Experiment 1: Blind Evaluation}\label{sec:method_exp1_analysis}
\paragraph{Setup.}
In Experiment 1, each model serves as a judge, evaluating all 300 selections without labels.
Each judge returns a JSON object with an integer score on a 1--7 scale for each dimension and a 2--3 sentence justification.
To average over stochastic variability in LLM scoring, each judge evaluates each selection three times, yielding $10 \text{ judges} \times 300 \text{ selections} \times 3 \text{ repetitions} = 9{,}000$ runs.
Within each run, both the order of the 20 constraints in the selection and the order of the four rubric dimensions are independently shuffled using a per-run seed, mitigating potential ordering biases.
The full evaluation prompt and response format are provided in \autoref{app:experiment_1}.

\paragraph{Raw self--other comparison.}
% confound 통제되지 않은 naive delta를 출발점으로 제시함. 
% 이 단계에서 나타나는 Self-Preference에 대한 해석은 보류. measurement confound 가능성을 다음 단락에서 검정한다는 식으로 연결하기.
For each judge, we compare the mean scores it assigns to its own selections versus those produced by other models, for the average score and each rubric dimension. 
To account for stochastic variability in LLM scoring, we test each judge's self-produced selections against its other-produced ones with Welch's two-sample t-tests; p-values are Holm-corrected across the ten judges.
This comparison controls for neither selector quality nor judge severity; it serves as the baseline against which the confound-controlled analysis is interpreted.

\paragraph{Confound-controlled estimation.}
% mixed-effects: score ~ is_self + (1|judge) + (1|selector) + (1|selection_id)
% 각 intercept가 통제하는 confound 명시해야 함.
% 검정 대상이 is_self임을 분명히 서술할 필요가 있다!!

To test whether self-preference remains once measurement confounds are controlled for, we fit a mixed-effects model:
\begin{equation*}
\small
\begin{aligned}
\texttt{score} = \beta_0\ & + \beta_S\cdot\texttt{is\_self} + (1\,|\,\texttt{judge})\\
                          & {}  + (1\,|\,\texttt{selector}) + (1\,|\,\texttt{selection\_id})
\end{aligned}
\end{equation*}

where $\beta_S$ is the coefficient of interest; the random intercepts absorb three measurement confounds: \texttt{judge} for rater severity, \texttt{selector} for selection quality, and \texttt{selection\_id} for non-independence across repeated evaluations of the same selection.
A positive, significant $\beta_S$ indicates that judges score their own selections higher after controls; a non-significant $\beta_S$ indicates that the apparent self-preference is attributable to the controlled confounds.

\subsubsection{Experiment 2: Labeled Evaluation}
\label{sec:method_exp2_analysis}
\paragraph{Setup.}
In Experiment~2, LLM judges evaluate selections under four label conditions: TL-self, TL-other, FL-self, and FL-other (see \autoref{tab:exp2_design}).
For each judge, we construct quality-matched pairs from Experiment~1 scores: a self- and an other-produced selection can be paired if the judge rated them within 0.25 on the overall mean and within 1 point on each dimension.
The thresholds ensure that paired selections are comparable not only in overall score but on each individual dimension, while producing enough pairs per judge for estimation.
Among eligible candidates, the closest match by total dimension distance is selected, yielding 40 pairs per judge (400 in total).
% On a 7-point scale, a within-1-point difference per dimension represents a practically negligible quality gap, and the 0.25 overall constraint ensures close aggregate matching.

Each evaluation run presents a single selection to the judge, so the attribution label affects an absolute rating rather than a direct pairwise comparison.
The label is inserted as an anonymous self/other phrase---``your own selection'' for self-labels and ``another language model'' for other-labels---without naming specific models.
This isolates the self-other distinction from model-identity effects (full prompts in \autoref{app:experiment_1}).
As in Experiment~1, each selection is evaluated three times, yielding $40 \text{ pairs} \times 10 \text{ judges} \times 4 \text{ conditions} \times 3 \text{ reps} = 4{,}800$ runs.

\newcolumntype{Y}{>{\centering\arraybackslash}X}

\begin{table}[t]
\centering
\small
\renewcommand{\arraystretch}{1.1}
\begin{tabularx}{0.9\columnwidth}{@{}l YY@{}}
\toprule
 & \textbf{Self label} & \textbf{Other label} \\
\midrule
 & \textbf{TL-self} & \textbf{FL-other} \\
\textbf{Self-produced}
  & {\footnotesize self selection + self label}
  & {\footnotesize self selection + other label} \\
\addlinespace
 & \textbf{FL-self} & \textbf{TL-other} \\
\textbf{Other-produced}
  & {\footnotesize other selection + self label}
  & {\footnotesize other selection + other label} \\
\bottomrule
\end{tabularx}
\caption{$2\times2$ design for Experiment~2, crossing the actual source of a selection with the label shown to the judge. TL = \emph{true label} (label matches the actual source); FL = \emph{false label} (label contradicts the actual source).}
\label{tab:exp2_design}
\end{table}

\paragraph{Label-induced attribution effect estimation.}
To investigate the authorship label effect and test whether it depends on the true content origin, we fit a mixed-effects model:
\begin{equation*}
\small
\begin{aligned}
\texttt{score} = \beta_0\ & + \beta_L\cdot\texttt{label} + \beta_A\cdot\texttt{actual} \\ & {}
                          + \beta_{LA}\cdot(\texttt{label}\times\texttt{actual}) \\
                          & {} + (1\,|\,\texttt{judge}) + (1\,|\,\texttt{selection\_id})
\end{aligned}
\end{equation*}

where \texttt{label} indicates the displayed source (self vs.\ other) and \texttt{actual} the true source, with random intercepts.
$\beta_L$ captures the main effect of the displayed label---the average score shift for a self- versus other-label---whereas $\beta_A$ reflects the main effect of actual source, the average score difference between self- and other-produced selections.
A non-significant $\beta_{LA}$ would indicate that the label effect is invariant across actual sources, with the same label-induced shift regardless of who actually produced the selection.

\paragraph{Directional decomposition.}
$\beta_L$ captures magnitude but not direction: it cannot distinguish whether self-labels inflate scores, other-labels deflate them, or both.
To resolve this, we use each judge's Experiment~1 (blind evaluation) scores as the baseline and compute the deviation of every labeled evaluation from this baseline.

\section{Results}
\subsection{Experiment 1: Apparent Self-Preference Disappears Under Control}

\subsubsection{Sanity Check}
\label{sec:sanity}
\begin{figure*}[t]
\centering
\includegraphics[width=0.95\textwidth]{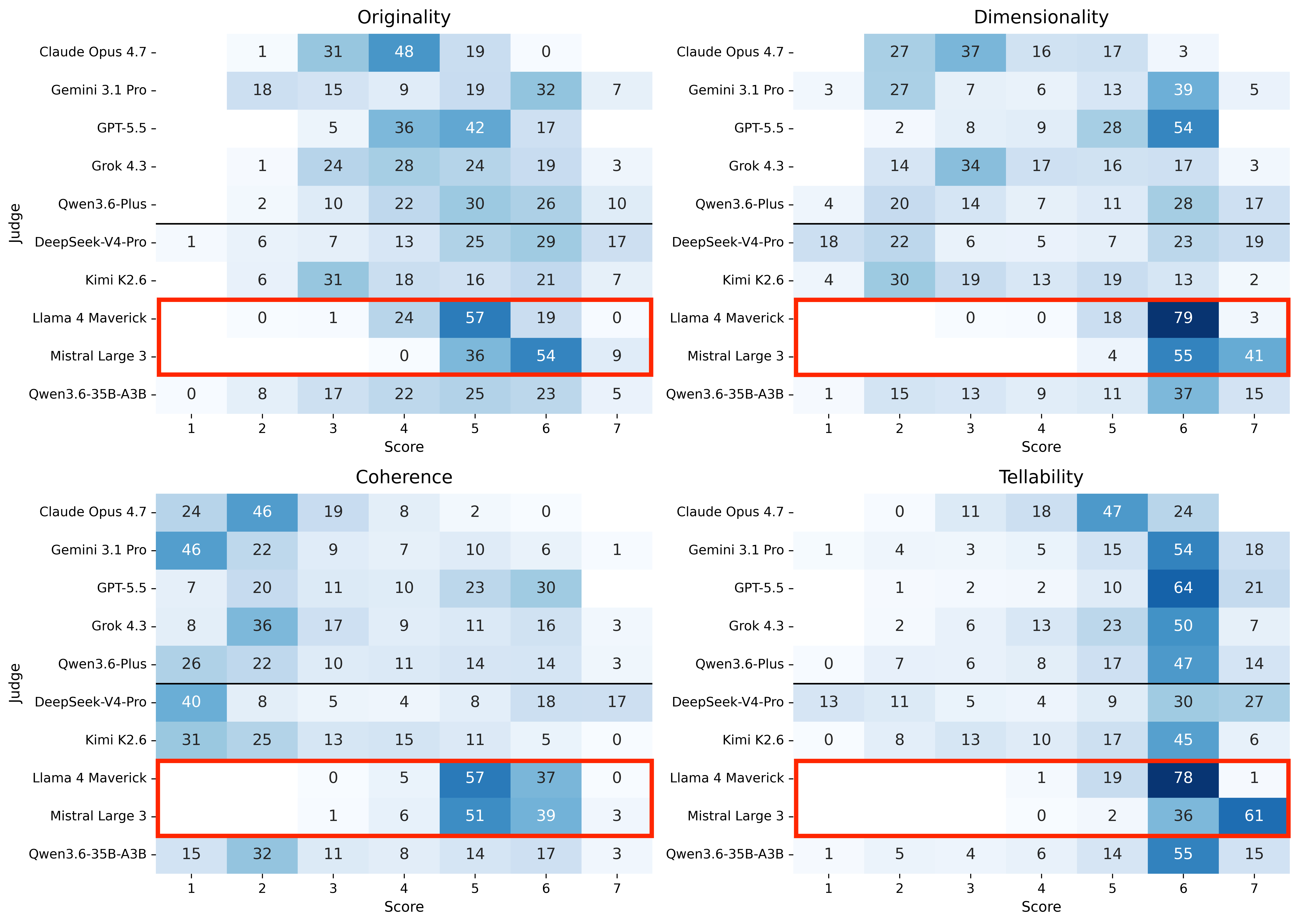}
\caption{Score frequency distribution across all Experiment~1 (blind) evaluations, used as a sanity check.
Each cell reports the percentage of evaluations in which a judge assigned that score for the given dimension.}
\label{fig:score_frequency}
\end{figure*}

In Experiment~1, eight of the ten judges produce meaningful score variation across selections (see \autoref{fig:score_frequency}).
The remaining two---Llama 4 Maverick and Mistral Large 3, both open-weight LLMs---fall short of the basic discriminative capacity required for self--other comparisons, placing at least 81\% of their ratings on two adjacent scores in every dimension.

\subsubsection{Raw Self-Other Comparison}

\begin{table}[t]
\centering
\small
\setlength{\tabcolsep}{6pt}
\renewcommand{\arraystretch}{1.1}
\begin{tabular}{@{}l rrrr@{}}
\toprule
\multicolumn{1}{c}{\textbf{Judge}} &
\multicolumn{1}{c}{\textbf{Self}} &
\multicolumn{1}{c}{\textbf{Other}} &
\multicolumn{1}{c}{\textbf{$\Delta$}} &
\multicolumn{1}{c}{\textbf{$p$}} \\
\midrule
Kimi K2.6        & 4.88 & 3.79 & $+1.09$ & $<.001$ \\
GPT-5.5          & 5.78 & 4.92 & $+0.86$ & $<.001$ \\
DeepSeek-V4-Pro  & 4.86 & 4.34 & $+0.52$ & $.350$ \\
Claude Opus 4.7  & 4.01 & 3.50 & $+0.51$ & $<.001$ \\
Qwen3.6-Plus     & 4.73 & 4.49 & $+0.24$ & $.484$ \\
Gemini 3.1 Pro   & 4.41 & 4.21 & $+0.21$ & $.484$ \\
Llama 4 Maverick & 5.51 & 5.47 & $+0.04$ & $.484$ \\
Mistral Large 3  & 5.94 & 6.02 & $-0.08$ & $.352$ \\
Grok 4.3         & 3.68 & 4.35 & $-0.67$ & $<.001$ \\
Qwen3.6-35B-A3B  & 3.75 & 4.65 & $-0.90$ & $<.001$ \\
\bottomrule
\end{tabular}
\caption{Raw self--other comparison on the average score, sorted by $\Delta = \text{Self}-\text{Other}$. $p$-values from Welch's two-sample t-tests, Holm-corrected across the ten judges. Confounds are not controlled; per-dimension results are in \autoref{app:raw_dims}.}
\label{tab:raw_self_other}
\end{table}

On average, seven of ten judges rate their own selections higher than others', but the significant gaps emerge in both directions: positive for Kimi K2.6, GPT-5.5, and Claude Opus 4.7, and negative for Grok 4.3 and Qwen3.6-35B-A3B (see \autoref{tab:raw_self_other}; per-dimension results in \autoref{app:raw_dims}).
However, the per-judge $\times$ selector matrix (see \autoref{fig:judge_selector_heatmap}) suggests this pattern reflects selection quality rather than genuine self-preference, as strong selectors receive higher scores from most judges, and weak selectors receive lower ones---including from themselves.
Kimi K2.6 illustrates this---it shows the largest raw self--other gap, yet its selections also rank among the highest from nearly every judge.
The raw gaps may therefore reflect producer-level quality differences rather than genuine self-preference, which we disentangle in the following analysis.

\subsubsection{Confound-Controlled Estimation}
Table~\ref{tab:confound} reports the self-preference coefficient ($\beta_S$) from \S\ref{sec:method_exp1_analysis} for the average score and each rubric dimension, before and after excluding the two low-discrimination judges identified in \S\ref{sec:sanity}.
Before the exclusion, $\beta_S$ is significantly positive on the average score and on \textit{Dimensionality}, \textit{Coherence}, and \textit{Tellability}, while \textit{Originality} shows a small negative effect.
After excluding the two, all positive effects disappear; only the negative \textit{Originality} effect remains, which is contrary to self-preference.
To ensure that the exclusion itself does not drive this result, we refit the model on all ten judges with a random slope for \textit{is\_self} by judge. 
We find no significant self-preference on any dimension, with substantial between-judge variability ($\tau$ = 0.41--0.63; see \autoref{app:random_slope}).
In sum, the raw self--other gap is attributable to the inflated ratings of the two low-discrimination judges and to selection quality rather than genuine self-preference.

\begin{table}[t]
\centering
\small
\setlength{\tabcolsep}{2.8pt}
\renewcommand{\arraystretch}{1.1}
\begin{tabular}{@{}l rc rc@{}}
\toprule
& \multicolumn{2}{c}{\textbf{Full (10 judges)}}
& \multicolumn{2}{c}{\textbf{Excl.\ low-discrim.\ (8)}} \\
\cmidrule(lr){2-3}\cmidrule(lr){4-5}
\textbf{Dim.} & \textbf{$\beta_S$} & \textbf{95\% CI}
              & \textbf{$\beta_S$} & \textbf{95\% CI} \\
\midrule
\textbf{Avg.} & $+0.181^{*}$ & [\,0.13,\,0.23\,]
              & $+0.003$     & [$-0.05$,\,0.06\,] \\
\midrule
Org. & $-0.144^{*}$ & [$-0.21$,\,$-0.08$] & $-0.133^{*}$ & [$-0.21$,\,$-0.05$] \\
Dim. & $+0.316^{*}$ & [\,0.24,\,0.39\,]   & $+0.039$     & [$-0.05$,\,0.12\,] \\
Coh. & $+0.276^{*}$ & [\,0.21,\,0.34\,]   & $+0.042$     & [$-0.03$,\,0.12\,] \\
Tel. & $+0.278^{*}$ & [\,0.21,\,0.34\,]   & $+0.065$     & [$-0.01$,\,0.14\,] \\
\bottomrule
\end{tabular}
\caption{Confound-controlled self-preference coefficient $\beta_S$ from the mixed-effects model, before and after excluding the two low-discrimination judges.
\textbf{Avg.}~= the average of the four rubric dimensions and the primary outcome; \textbf{Org.}~= 
\textit{Originality}, \textbf{Dim.}~= \textit{Dimensionality}, \textbf{Coh.}~= \textit{Coherence}, \textbf{Tel.}~= \textit{Tellability}.
$^{*}p < .05$.
The positive effects on the average score and three dimensions vanish once low-discrimination judges are removed; only a \emph{negative} \textit{Originality} effect remains, which runs counter to self-preference (though not robust to random-slope specification; \autoref{app:random_slope}).}
\label{tab:confound}
\end{table}

% 4.2.
\subsection{Experiment 2: Labels Alone Induce Bidirectional Attribution Bias}
\subsubsection{Per-Condition Descriptive Statistics}

\autoref{tab:exp2_desc} reports descriptive scores across four label–source conditions, illustrating how self- versus other-attribution shifts evaluations of identical selections regardless of actual authorship.
Within each \textit{Actual} block, self-labels elevate scores over other-labels by 0.29--0.57 across all four dimensions without exception---notably including \textit{Originality}, which had shown, if anything, a weak self-deprecation tendency in the blind condition.
By contrast, comparing across actual sources at the same label yields near-identical scores (4.92 vs.\ 4.91 average under self-labels; 4.50 vs.\ 4.48 under other-labels), suggesting that the label, not the content, drives the shift in this experiment.

\begin{table*}[t]
\centering
\small
\setlength{\tabcolsep}{4pt}
\renewcommand{\arraystretch}{1.1}
\begin{tabular}{@{}>{\centering\arraybackslash}m{1.8cm} l cccccc@{}}
\toprule
\multicolumn{1}{c}{\textbf{Actual}} & \textbf{Condition} & \multicolumn{1}{c}{\textbf{Average}} & \multicolumn{1}{c}{\textbf{Org.}} & \multicolumn{1}{c}{\textbf{Dim.}} & \multicolumn{1}{c}{\textbf{Coh.}} & \multicolumn{1}{c}{\textbf{Tel.}} \\
\midrule
Self & TL-self  & 4.92 (1.04) & 4.76 (1.24) & 5.16 (1.55) & 3.85 (1.74) & 5.91 (1.09) \\
-produced  & FL-other & 4.50 (1.10) & 4.28 (1.25) & 4.60 (1.60) & 3.56 (1.69) & 5.55 (1.14) \\
\midrule
Other & FL-self  & 4.91 (1.06) & 4.83 (1.23) & 5.09 (1.57) & 3.92 (1.80) & 5.79 (1.12) \\
-produced & TL-other & 4.48 (1.16) & 4.30 (1.19) & 4.52 (1.68) & 3.63 (1.82) & 5.47 (1.22) \\
\bottomrule
\end{tabular}
\caption{Per-condition descriptive statistics for Experiment~2. Mean (SD) across all judges, pairs, and repetitions.
TL = label matches actual source; FL = label contradicts actual source.}
\label{tab:exp2_desc}
\end{table*}

\subsubsection{Label-Induced Attribution Effect}
\autoref{tab:exp2_mixed} presents the mixed-effects model estimates from \S\ref{sec:method_exp2_analysis}.
The label main effect ($\beta_L$) is significantly positive across all four dimensions and for the average score, while neither the actual-source effect ($\beta_A$) nor the label$\times$actual interaction ($\beta_{LA}$) reaches significance on any dimension.
As a manipulation check, $\beta_A$—the effect of the selection's true source—should be near zero if quality matching succeeded; its non-significance on every dimension indicates that matching left no substantial content-driven differences, so the displayed-label effect ($\beta_L$) cannot be attributed to residual quality gaps. 
%The non-significant $\beta_A$ indicates that quality matching successfully removed content-driven differences; the non-significant $\beta_{LA}$ further confirms that the label effect operates independently of the true source, ruling out self-recognition \cite{panickssery2024llm} as the driver.
Together, these results show that the displayed attribution label alone can shift scores, even when the evaluation target is identical.

\begin{table}[t]
\centering
\renewcommand{\arraystretch}{1.1}
\small
\setlength{\tabcolsep}{4pt}
\begin{tabular}{@{}ccccc@{}}
\toprule
\textbf{Dim.} & \textbf{Effect} & \textbf{Estimate} & \textbf{95\% CI} & $\mathbf{p}$ \\
\midrule
             & $\beta_L$    & $\mathbf{+0.43^{***}}$ & $[+0.39, +0.47]$ & $<.001$ \\
\textbf{Avg.} & $\beta_A$    & $+0.02$            & $[-0.09, +0.14]$ & $.689$  \\
              & $\beta_{LA}$ & $-0.01$            & $[-0.06, +0.05]$ & $.823$  \\
\midrule
     & $\beta_L$    & $\mathbf{+0.53^{***}}$ & $[+0.47, +0.60]$ & $<.001$ \\
Org. & $\beta_A$    & $-0.03$            & $[-0.15, +0.09]$ & $.663$  \\
     & $\beta_{LA}$ & $-0.05$            & $[-0.14, +0.05]$ & $.346$  \\
\midrule
     & $\beta_L$    & $\mathbf{+0.57^{***}}$ & $[+0.50, +0.65]$ & $<.001$ \\
Dim. & $\beta_A$    & $+0.11$            & $[-0.06, +0.27]$ & $.197$  \\
     & $\beta_{LA}$ & $-0.02$            & $[-0.12, +0.08]$ & $.721$  \\
\midrule
     & $\beta_L$    & $\mathbf{+0.29^{***}}$ & $[+0.24, +0.35]$ & $<.001$ \\
Coh. & $\beta_A$    & $-0.06$            & $[-0.26, +0.14]$ & $.541$  \\
     & $\beta_{LA}$ & $-0.01$            & $[-0.09, +0.07]$ & $.805$  \\
\midrule
     & $\beta_L$    & $\mathbf{+0.32^{***}}$ & $[+0.26, +0.38]$ & $<.001$ \\
Tel. & $\beta_A$    & $+0.07$            & $[-0.05, +0.20]$ & $.257$  \\
     & $\beta_{LA}$ & $+0.05$            & $[-0.04, +0.13]$ & $.261$  \\
\bottomrule
\end{tabular}
\caption{Fixed-effect estimates from the mixed-effects models for Experiment~2. 
$\beta_L$ captures the displayed-label effect; $\beta_A$ the actual-source effect (expected $\approx 0$ under successful quality matching); $\beta_{LA}$ their interaction. $^{***} p < .001$. 
Across all dimensions, $\beta_L$ is the only significant fixed effect, and this pattern is robust to excluding two low-discrimination judges (see \autoref{app:exp2_mixed_excl}).}
\label{tab:exp2_mixed}
\end{table}

\subsubsection{A Bidirectional Label-Induced Attribution Bias}
Table~\ref{tab:exp2_deviation} indicates that the label-induced attribution bias operates bidirectionally.
All ten judges rate self-labels above other-labels, and six follow the dominant \textit{symmetric} pattern---inflating scores under self-labels while simultaneously deflating them under other-labels.
Three exhibit one-sided variants: Claude Opus 4.7 and DeepSeek-V4-Pro show significant inflation under self-labels but no reliable change under other-labels (\textit{self-boost}), while GPT-5.5 shows the reverse (\textit{other-penalty}).
Llama 4 Maverick deflates under both labels (\textit{general-penalty}), consistent with its low-discrimination behavior in the sanity check. 
These findings refine the previous analysis by showing that the label main effect ($\beta_L$) operates in two directions---self-inflation and other-deflation---that may co-occur or appear alone.
%arises from two distinct mechanisms---self-preference and other-derogation---that may co-occur or appear alone.

\begin{table}[t]
\centering
\renewcommand{\arraystretch}{1.1}
\small
\setlength{\tabcolsep}{2pt}
\begin{tabular}{@{}lccc@{}}
\toprule
\multicolumn{1}{c}{\textbf{Judge}} & \textbf{Self-label} & \textbf{Other-label} & \textbf{Pattern} \\
\midrule
Qwen3.6-35B-A3B   & $+0.131^*$ & $-0.523^*$  & symmetric       \\
Gemini 3.1 Pro    & $+0.302^*$ & $-0.339^*$  & symmetric       \\
Qwen3.6-Plus      & $+0.175^*$ & $-0.444^*$  & symmetric       \\
Kimi K2.6         & $+0.147^*$ & $-0.386^*$  & symmetric       \\
Mistral Large 3   & $+0.081^*$ & $-0.265^*$  & symmetric       \\
Grok 4.3          & $+0.126^*$ & $-0.195^*$  & symmetric       \\
\midrule[\lightrulewidth]
DeepSeek-V4-Pro   & $+0.443^*$ & $+0.101$    & self-boost      \\
Claude Opus 4.7   & $+0.275^*$ & $-0.023$    & self-boost      \\
\midrule[\lightrulewidth]
GPT-5.5           & $-0.019$   & $-0.340^*$  & other-penalty   \\
\midrule[\lightrulewidth]
Llama 4 Maverick  & $-0.192^*$ & $-0.384^*$  & general-penalty \\
\bottomrule
\end{tabular}
\caption{Per-judge mean deviation from the Experiment~1 (no-label) baseline under self- and other-label conditions on the average score. Patterns: \textit{symmetric} (self $\uparrow$, other $\downarrow$), \textit{self-boost} (self $\uparrow$ only), \textit{other-penalty} (other $\downarrow$ only), \textit{general-penalty} (both $\downarrow$). Rows grouped by pattern; within each group, judges are sorted by the magnitude of the label-induced shift. $^* p < .05$ (one-sample $t$-test, H$_0$: $\mu = 0$).}
\label{tab:exp2_deviation}
\end{table}

\begin{figure*}
    \centering
    \includegraphics[width=0.72\linewidth]{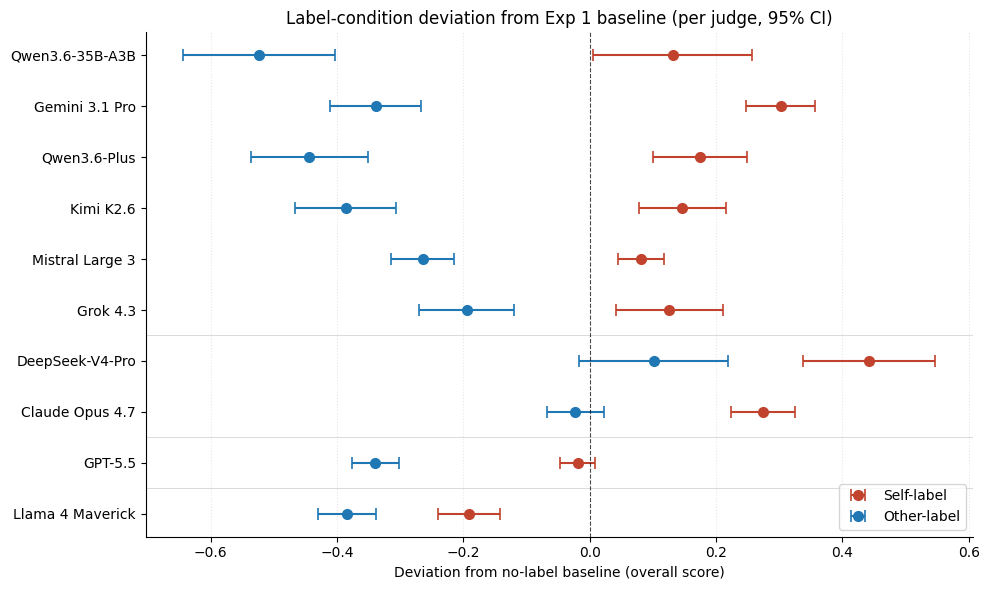}
    \caption{Per-judge mean deviation from the Experiment 1 (blind evaluation) baseline under self-labels (red) and other-labels (blue) on the average score, with 95\% CIs.}
    \label{fig:placeholder}
\end{figure*}

\section{Discussion}
% 1. 결과 해석
\subsection{Candidate Mechanisms: Familiarity vs. Self-Recognition}
Across our two experiments, apparent self-preference disappears under blind evaluation once confounds are controlled, and gives way to a label-driven shift when self- and other-labels are attached to identical targets.
If LLM judges favor lower-perplexity outputs, as \citet{wataoka2024self} argue, self-preference should have arisen in Experiment~1: each judge exhibits a distinct selection pattern, so its own selections should be the most familiar to it even though model-specific stylistic cues are structurally eliminated. 
However, self-preference is no longer significant once selection quality and judge severity are controlled.
More fundamentally, the familiarity hypothesis cannot explain why attributed authorship shifts evaluation scores when the targets themselves remain identical.
Familiarity thus does not appear to be a substantial driver of self-preference in our setting; the higher mean scores that judges initially assigned to their own selections are better attributed to measurement confounds.

By contrast, the self-recognition hypothesis of \citet{panickssery2024llm}, who reported a linear correlation between self-recognition ability and self-preference strength, provides a useful starting point for interpreting both experiments.
We view Experiment~1 as a setting where self-recognition is hindered by the nature of the evaluation targets: although each model's selections carry a statistically recoverable signature (\S\ref{sec:selection_task}), this signal may be too weak for a judge to recognize its own selections during evaluation.
In Experiment~2, in contrast, we speculate that explicit self- and other-labels establish the conditions under which the self/other distinction becomes operative.
While assigning explicit labels differs from text-based implicit self-recognition, our findings align with the account of \citet{panickssery2024llm} in a broader sense: what matters is whether authorship is marked as self or other, regardless of whether it is inferred through self-recognition or supplied by an explicit label.
Still, how LLM judges internally represent these self/other attributions warrants further examination.

\subsection{Generalizability and Broader Implications}
\label{sec:Dis2}
% BROADER IMPLICATION
In our two experiments, the evaluation targets are narrative selections rather than generated texts.
While this design separates content from model-specific surface cues and thus enables rigorous control of potential confounds, our main findings are established on this task.
To examine whether the observed label-induced effects extend beyond the constraint-based narrative selection setting, we conducted a supplementary replication applying the Experiment~2 design to the evaluation of full-length generated research proposals, following \citet{si2025can}.

Specifically, the ten models each generated three research proposals for each of three topics, yielding 90 proposals in total.
We then normalized the style of all proposals using the style-normalization prompt of \citet{si2025can} with a fixed non-judge model (\textit{Claude Haiku 4.5}) to reduce stylistic variation across models. 
All ten judges subsequently evaluated all 90 normalized proposals in a blind setting using the accompanying review form verbatim, with anchored 1--10 scales for \textit{Novelty}, \textit{Feasibility}, \textit{Expected Effectiveness}, \textit{Excitement}, and \textit{Overall}.
For the labeled evaluation, we constructed 85 self--other pairs from proposals whose blind \textit{Overall} scores differed by no more than 0.5. 
Each proposal was then evaluated three times under both self and other labels in a 2$\times$2 design using the same review form, with anonymous self/other phrases as in our main experiments.

\begin{table}[t]
\small
\centering
\begin{tabular}{llrr}
\toprule
Dim. & Effect & Estimate & $p$ \\
\midrule
 & $\beta_L$ & $+0.196^{***}$ & $<.001$ \\
Overall & $\beta_A$ & $+0.011$ & $.897$ \\
 & $\beta_{LA}$ & $+0.016$ & $.779$ \\
\midrule
 & $\beta_L$ & $+0.244^{**}$ & $.003$ \\
Nov. & $\beta_A$ & $-0.014$ & $.908$ \\
 & $\beta_{LA}$ & $-0.043$ & $.426$ \\
\midrule
 & $\beta_L$ & $-0.116^{*}$ & $.016$ \\
Fea. & $\beta_A$ & $+0.125$ & $.182$ \\
 & $\beta_{LA}$ & $+0.051$ & $.366$ \\
\midrule
 & $\beta_L$ & $+0.047$ & $.202$ \\
Eff. & $\beta_A$ & $+0.002$ & $.989$ \\
 & $\beta_{LA}$ & $+0.078$ & $.133$ \\
\midrule
 & $\beta_L$ & $+0.206^{***}$ & $<.001$ \\
Exc. & $\beta_A$ & $+0.019$ & $.810$ \\
 & $\beta_{LA}$ & $-0.008$ & $.880$ \\
\bottomrule
\end{tabular}
\caption{Fixed-effect estimates from the mixed-effects models for the
research-proposal replication following \citet{si2025can}.
\textbf{Nov.} = \textit{Novelty}, \textbf{Fea.} = \textit{Feasibility},
\textbf{Eff.} = \textit{Expected Effectiveness}, \textbf{Exc.} =
\textit{Excitement}. $\beta_L$ captures the displayed-label effect;
$\beta_A$ the actual-source effect; $\beta_{LA}$ their interaction.
$^{*}p<.05$, $^{**}p<.01$, $^{***}p<.001$.}
\label{tab:proposal_label_effect}
\end{table}

The results in Table \ref{tab:proposal_label_effect} show that label attribution significantly shifted evaluations of identical research proposals, while neither actual authorship nor the interaction between label and actual authorship was significant.
The effect was dimension-specific, increasing \textit{Overall}, \textit{Novelty}, and \textit{Excitement} scores while decreasing \textit{Feasibility} scores. 
The reversal for \textit{Feasibility} indicates that attribution does not simply induce uniformly more favorable scoring; the direction of the shift varies with the evaluation dimension.
The recurrence of systematic attribution effects suggests that the bias is not specific to the narrative-selection setup and may extend to more naturalistic evaluation settings.

These findings have two implications for the measurement and deployment of LLM judges.
First, self-preference measured without sophisticated confound controls is likely  overstated.
The raw self--other gaps in our data were attributable to selection quality and judge severity, so future measurement should identify and control for such confounds. 
Second, our results identify a source of evaluation bias operating at the level of assigned authorship, independently of content.
If an evaluation pipeline marks outputs as the judge's own by design---as in self-refinement, where self-preference is known to amplify  \citep{xu2024pride}---such attribution may distort the resulting evaluations. 
Future research is needed to determine if withholding or obfuscating attribution reliably suppresses this bias across evaluation settings.

\section{Conclusion}
This study asks whether self-preference in LLM judges persists once surface-level cues and confounders are controlled, and whether a self-authorship label alone can induce such bias under the same controls.
Drawing on a narrative selection task that eliminates stylistic features by design, we examine LLM judge behavior under two settings.
Our results show that, while most LLM judges assign higher mean scores to their own selections in blind evaluation, this apparent self-preference disappears once selection quality and judge severity are controlled.
When authorship labels are assigned, however, they systematically shift judges' evaluations regardless of the evaluation target's actual source: the most common pattern is a symmetric shift---judges inflate scores under self-labels and deflate them under other-labels---while some models show one-sided variants, inflating self-labeled or deflating other-labeled scores alone.
A supplementary replication on the evaluation of generated research proposals suggests that this label-induced bias extends beyond the narrative selection setting, though its direction varies by dimension.
Our contributions are twofold.
First, by introducing an experimental design that structurally eliminates surface confounders, we provide direct evidence that self/other labels alone can induce bidirectional evaluation bias in LLM judges.
Second, we suggest that open-ended, ground-truth-free tasks can serve as controlled instruments for studying LLM judge behavior.

\section*{Limitations}
Our main experiments are confined to a single creative-selection  domain---a setting that prioritizes rigorous experimental control over ecological validity.
The replication on research-proposal evaluation in \S\ref{sec:Dis2} provides initial evidence of generalization, but to one additional domain only.
Whether similar effects emerge in settings with verifiable ground truth, such as reasoning or instruction-following evaluation, remains unclear.
In addition, all experiments in this study are conducted in English, so the cross-lingual generalizability of the label-induced bias remains an open question.
Moreover, our labeled evaluations---in both the main and the additional experiments---rely on explicit self/other attribution prompts, whereas in real-world LLM-as-a-judge deployments authorship information is often implicit or unavailable.
It remains unexamined whether real-world attribution cues, such as stylistic fingerprints, model-specific linguistic patterns, or interface-level signals, induce the same bias.
Finally, the 4-axis rubric we establish is not validated against human evaluation. 
It serves as a structured interpretive frame that constrains how judges interpret the evaluation criteria, rather than an absolute measure of quality.

\bibliography{custom}

@inproceedings{zheng2023judging,
  title     = {Judging {LLM}-as-a-Judge with {MT-Bench} and {Chatbot Arena}},
   author = {Zheng, Lianmin and Chiang, Wei-Lin and Sheng, Ying and Zhuang, Siyuan and Wu, Zhanghao and Zhuang, Yonghao and Lin, Zi and Li, Zhuohan and Li, Dacheng and Xing, Eric and Zhang, Hao and Gonzalez, Joseph and Stoica, Ion},
 booktitle = {Advances in Neural Information Processing Systems},
 editor = {A. Oh and T. Naumann and A. Globerson and K. Saenko and M. Hardt and S. Levine},
 pages = {46595--46623},
 publisher = {Curran Associates, Inc.},
 url = {https://dl.acm.org/doi/10.5555/3666122.3668142},
 volume = {36},
 year = {2023}
}

@article{chen2025llm,
  title   = {Do {LLM} Evaluators Prefer Themselves for a Reason?},
  author  = {Chen, Wei-Lin and Wei, Zhepei and Zhu, Xinyu and Feng, Shi and Meng, Yu},
  journal = {arXiv preprint arXiv:2504.03846},
  year    = {2025},
  doi     = {10.48550/arXiv.2504.03846},
  url     = {https://arxiv.org/abs/2504.03846}
}

@inproceedings{chakrabarty2024art,
  title     = {Art or Artifice? Large Language Models and the False Promise of Creativity},
  author    = {Chakrabarty, Tuhin and Laban, Philippe and Agarwal, Divyansh and Muresan, Smaranda and Wu, Chien-Sheng},
  booktitle = {Proceedings of the 2024 CHI Conference on Human Factors in Computing Systems},
  year      = {2024},
  pages     = {1--34},
  address   = {New York, NY, USA},
  publisher = {Association for Computing Machinery},
  doi       = {10.1145/3613904.3642731},
  url       = {https://doi.org/10.1145/3613904.3642731}
}

@inproceedings{li2025generation,
  title     = {From Generation to Judgment: Opportunities and Challenges of {LLM}-as-a-judge},
  author    = {Li, Dawei and Jiang, Bohan and Huang, Liangjie and Beigi, Alimohammad and Zhao, Chengshuai and Tan, Zhen and Bhattacharjee, Amrita and Jiang, Yuxuan and Chen, Canyu and Wu, Tianhao and Shu, Kai and Cheng, Lu and Liu, Huan},
  booktitle = {Proceedings of the 2025 Conference on Empirical Methods in Natural Language Processing},
  month     = nov,
  year      = {2025},
  address   = {Suzhou, China},
  publisher = {Association for Computational Linguistics},
  pages     = {2757--2791},
  doi       = {10.18653/v1/2025.emnlp-main.138},
  url       = {https://aclanthology.org/2025.emnlp-main.138/},
  isbn      = {979-8-89176-332-6}
}

@inproceedings{liu2023g,
  title     = {{G}-Eval: {NLG} Evaluation using {GPT}-4 with Better Human Alignment},
  author    = {Liu, Yang and Iter, Dan and Xu, Yichong and Wang, Shuohang and Xu, Ruochen and Zhu, Chenguang},
  booktitle = {Proceedings of the 2023 Conference on Empirical Methods in Natural Language Processing},
  month     = dec,
  year      = {2023},
  address   = {Singapore},
  publisher = {Association for Computational Linguistics},
  pages     = {2511--2522},
  doi       = {10.18653/v1/2023.emnlp-main.153},
  url       = {https://aclanthology.org/2023.emnlp-main.153/}
}

@inproceedings{bai2023benchmarking,
  title     = {Benchmarking Foundation Models with Language-Model-as-an-Examiner},
  author    = {Bai, Yushi and Ying, Jiahao and Cao, Yixin and Lv, Xin and He, Yuze and Wang, Xiaozhi and Yu, Jifan and Zeng, Kaisheng and Xiao, Yijia and Lyu, Haozhe and Zhang, Jiayin and Li, Juanzi and Hou, Lei},
  booktitle = {Advances in Neural Information Processing Systems},
  volume    = {36},
  pages     = {78142--78167},
  year      = {2023},
  url       = {https://proceedings.neurips.cc/paper_files/paper/2023/hash/f64e55d03e2fe61aa4114e49cb654acb-Abstract-Datasets_and_Benchmarks.html}
}

@inproceedings{koo2024benchmarking,
  title     = {Benchmarking Cognitive Biases in Large Language Models as Evaluators},
  author    = {Koo, Ryan and Lee, Minhwa and Raheja, Vipul and Park, Jong Inn and Kim, Zae Myung and Kang, Dongyeop},
  booktitle = {Findings of the Association for Computational Linguistics: ACL 2024},
  month     = aug,
  year      = {2024},
  address   = {Bangkok, Thailand},
  publisher = {Association for Computational Linguistics},
  pages     = {517--545},
  doi       = {10.18653/v1/2024.findings-acl.29},
  url       = {https://aclanthology.org/2024.findings-acl.29/}
}

@inproceedings{ye2025justice,
  title     = {Justice or {P}rejudice? {Q}uantifying {B}iases in {LLM}-as-a-{J}udge},
  author    = {Ye, Jiayi and Wang, Yanbo and Huang, Yue and Chen, Dongping and Zhang, Qihui and Moniz, Nuno and Gao, Tian and Geyer, Werner and Huang, Chao and Chen, Pin-Yu and Chawla, Nitesh and Zhang, Xiangliang},
  booktitle = {International Conference on Learning Representations},
  year      = {2025},
  url       = {https://proceedings.iclr.cc/paper_files/paper/2025/hash/fdca08d371e4b6c031397909e20043bd-Abstract-Conference.html}
}

@inproceedings{wang2024large,
  title     = {Large {L}anguage {M}odels are not {F}air {E}valuators},
  author    = {Wang, Peiyi and Li, Lei and Chen, Liang and Cai, Zefan and Zhu, Dawei and Lin, Binghuai and Cao, Yunbo and Kong, Lingpeng and Liu, Qi and Liu, Tianyu and Sui, Zhifang},
  booktitle = {Proceedings of the 62nd Annual Meeting of the Association for Computational Linguistics (Volume 1: Long Papers)},
  month     = aug,
  year      = {2024},
  address   = {Bangkok, Thailand},
  publisher = {Association for Computational Linguistics},
  pages     = {9440--9450},
  doi       = {10.18653/v1/2024.acl-long.511},
  url       = {https://aclanthology.org/2024.acl-long.511/}
}

@article{saunders2022self,
  title   = {Self-critiquing Models for Assisting Human Evaluators},
  author  = {Saunders, William and Yeh, Catherine and Wu, Jeff and Bills, Steven and Ouyang, Long and Ward, Jonathan and Leike, Jan},
  journal = {arXiv preprint arXiv:2206.05802},
  year    = {2022},
  doi     = {10.48550/arXiv.2206.05802},
  url     = {https://arxiv.org/abs/2206.05802}
}

@article{wataoka2024self,
  title   = {Self-Preference Bias in {LLM}-as-a-Judge},
  author  = {Wataoka, Koki and Takahashi, Tsubasa and Ri, Ryokan},
  journal = {arXiv preprint arXiv:2410.21819},
  year    = {2024},
  doi     = {10.48550/arXiv.2410.21819},
  url     = {https://arxiv.org/abs/2410.21819}
}

@article{pombal2026self,
  title   = {Self-Preference Bias in Rubric-Based Evaluation of Large Language Models},
  author  = {Pombal, Jos{\'e} and Rei, Ricardo and Martins, Andr{\'e} F. T.},
  journal = {arXiv preprint arXiv:2604.06996},
  year    = {2026},
  doi     = {10.48550/arXiv.2604.06996},
  url     = {https://arxiv.org/abs/2604.06996}
}

@inproceedings{chen2025beyond,
  title     = {Beyond the Surface: Measuring Self-Preference in {LLM} Judgments},
  author    = {Chen, Zhi-Yuan and Wang, Hao and Zhang, Xinyu and Hu, Enrui and Lin, Yankai},
  booktitle = {Proceedings of the 2025 Conference on Empirical Methods in Natural Language Processing},
  month     = nov,
  year      = {2025},
  address   = {Suzhou, China},
  publisher = {Association for Computational Linguistics},
  pages     = {1653--1672},
  doi       = {10.18653/v1/2025.emnlp-main.86},
  url       = {https://aclanthology.org/2025.emnlp-main.86/},
  isbn      = {979-8-89176-332-6}
}

@inproceedings{mahbub2026mitigating,
  title     = {Mitigating Self-Preference by Authorship Obfuscation},
  author    = {Mahbub, Taslim and Feng, Shi},
  booktitle = {Proceedings of the AAAI Conference on Artificial Intelligence},
  volume    = {40},
  number    = {44},
  pages     = {37701--37708},
  year      = {2026},
  doi       = {10.1609/aaai.v40i44.41105},
  url       = {https://ojs.aaai.org/index.php/AAAI/article/view/41105}
}

@inproceedings{liu2024llms,
  title     = {{LLM}s as Narcissistic Evaluators: When Ego Inflates Evaluation Scores},
  author    = {Liu, Yiqi and Moosavi, Nafise and Lin, Chenghua},
  booktitle = {Findings of the Association for Computational Linguistics: ACL 2024},
  month     = aug,
  year      = {2024},
  address   = {Bangkok, Thailand},
  publisher = {Association for Computational Linguistics},
  pages     = {12688--12701},
  doi       = {10.18653/v1/2024.findings-acl.753},
  url       = {https://aclanthology.org/2024.findings-acl.753/}
}

@inproceedings{shi2025judging,
  title     = {Judging the Judges: A Systematic Study of Position Bias in {LLM}-as-a-Judge},
  author    = {Shi, Lin and Ma, Chiyu and Liang, Wenhua and Diao, Xingjian and Ma, Weicheng and Vosoughi, Soroush},
  booktitle = {Proceedings of the 14th International Joint Conference on Natural Language Processing and the 4th Conference of the Asia-Pacific Chapter of the Association for Computational Linguistics},
  month     = dec,
  year      = {2025},
  address   = {Mumbai, India},
  publisher = {The Asian Federation of Natural Language Processing and The Association for Computational Linguistics},
  pages     = {292--314},
  doi       = {10.18653/v1/2025.ijcnlp-long.18},
  url       = {https://aclanthology.org/2025.ijcnlp-long.18/},
  isbn      = {979-8-89176-298-5}
}

@inproceedings{panickssery2024llm,
  title     = {{LLM} Evaluators Recognize and Favor Their Own Generations},
  author    = {Panickssery, Arjun and Bowman, Samuel R. and Feng, Shi},
  booktitle = {Advances in Neural Information Processing Systems},
  volume    = {37},
  pages     = {68772--68802},
  year      = {2024},
  url       = {https://proceedings.neurips.cc/paper_files/paper/2024/hash/7f1f0218e45f5414c79c0679633e47bc-Abstract-Conference.html}
}

@inproceedings{xu2024pride,
  title     = {Pride and Prejudice: {LLM} Amplifies Self-Bias in Self-Refinement},
  author    = {Xu, Wenda and Zhu, Guanglei and Zhao, Xuandong and Pan, Liangming and Li, Lei and Wang, William},
  booktitle = {Proceedings of the 62nd Annual Meeting of the Association for Computational Linguistics (Volume 1: Long Papers)},
  month     = aug,
  year      = {2024},
  address   = {Bangkok, Thailand},
  publisher = {Association for Computational Linguistics},
  pages     = {15474--15492},
  doi       = {10.18653/v1/2024.acl-long.826},
  url       = {https://aclanthology.org/2024.acl-long.826/}
}

@article{roytburg2026llm,
  title   = {Are {LLM} Evaluators Really Narcissists? Sanity Checking Self-Preference Evaluations},
  author  = {Roytburg, Dani and Bozoukov, Matthew and Nguyen, Matthew and Barzdukas, Jou and Puig-Hall, Mackenzie and Oozeer, Narmeen},
  journal = {arXiv preprint arXiv:2601.22548},
  year    = {2026},
  doi     = {10.48550/arXiv.2601.22548},
  url     = {https://arxiv.org/abs/2601.22548}
}

@inproceedings{sun2026label,
    title = "Label Effects: Shared Heuristic Reliance in Trust Assessment by Humans and {LLM}-as-a-Judge",
    author = "Sun, Xin  and
      Wu, Di  and
      Qin, Sijing  and
      Echizen, Isao  and
      El Ali, Abdallah  and
      Sugawara, Saku",
    editor = "Liakata, Maria  and
      Moreira, Viviane P.  and
      Zhang, Jiajun  and
      Jurgens, David",
    booktitle = "Proceedings of the 64th Annual Meeting of the {A}ssociation for {C}omputational {L}inguistics (Volume 1: Long Papers)",
    month = jul,
    year = "2026",
    address = "San Diego, California, United States",
    publisher = "Association for Computational Linguistics",
    url = "https://aclanthology.org/2026.acl-long.1495/",
    doi = "10.18653/v1/2026.acl-long.1495",
    pages = "32378--32392",
    ISBN = "979-8-89176-390-6",
}

@article{marioriyad2025silent,
  title   = {The Silent Judge: Unacknowledged Shortcut Bias in {LLM}-as-a-Judge},
  author  = {Marioriyad, Arash and Rohban, Mohammad Hossein and Baghshah, Mahdieh Soleymani},
  journal = {arXiv preprint arXiv:2509.26072},
  year    = {2025},
  doi     = {10.48550/arXiv.2509.26072},
  url     = {https://arxiv.org/abs/2509.26072}
}

@article{gu2026survey,
  title   = {A Survey on {LLM}-as-a-Judge},
  author  = {Gu, Jiawei and Jiang, Xuhui and Shi, Zhichao and Tan, Hexiang and Zhai, Xuehao and Xu, Chengjin and Li, Wei and Shen, Yinghan and Ma, Shengjie and Liu, Honghao and Wang, Saizhuo and Zhang, Kun and Lin, Zhouchi and Zhang, Bowen and Ni, Lionel and Gao, Wen and Wang, Yuanzhuo and Guo, Jian},
  journal = {The Innovation},
  volume  = {7},
  number  = {6},
  pages   = {101253},
  year    = {2026},
  doi     = {10.1016/j.xinn.2025.101253},
  url     = {https://doi.org/10.1016/j.xinn.2025.101253}
}

@inproceedings{fein2026litbench,
  title     = {{L}it{B}ench: A Benchmark and Dataset for Reliable Evaluation of Creative Writing},
  author    = {Fein, Daniel and Russo, Sebastian and Xiang, Violet and Jolly, Kabir and Rafailov, Rafael and Haber, Nick},
  booktitle = {Proceedings of the 19th Conference of the European Chapter of the Association for Computational Linguistics (Volume 1: Long Papers)},
  month     = mar,
  year      = {2026},
  address   = {Rabat, Morocco},
  publisher = {Association for Computational Linguistics},
  pages     = {7740--7755},
  doi       = {10.18653/v1/2026.eacl-long.362},
  url       = {https://aclanthology.org/2026.eacl-long.362/},
  isbn      = {979-8-89176-380-7}
}

@article{saraf2025quantifying,
  title   = {Quantifying Label-Induced Bias in Large Language Model Self- and Cross-Evaluations},
  author  = {Saraf, Muskan and Boroujeni, Sajjad Rezvani and Beaudry, Justin and Abedi, Hossein and Bush, Tom},
  journal = {arXiv preprint arXiv:2508.21164},
  year    = {2025},
  doi     = {10.48550/arXiv.2508.21164},
  url     = {https://arxiv.org/abs/2508.21164}
}

@inproceedings{jung2026style,
    title = "Style over Story: Measuring {LLM} Narrative Preferences via Structured Selection",
    author = "Jung, Donghoon  and
      Choi, Jiwoo  and
      Chae, Songeun  and
      Jung, Seohyon",
    editor = "Liakata, Maria  and
      Moreira, Viviane P.  and
      Zhang, Jiajun  and
      Jurgens, David",
    booktitle = "Findings of the {A}ssociation for {C}omputational {L}inguistics: {ACL} 2026",
    month = jul,
    year = "2026",
    address = "San Diego, California, United States",
    publisher = "Association for Computational Linguistics",
    url = "https://aclanthology.org/2026.findings-acl.1361/",
    doi = "10.18653/v1/2026.findings-acl.1361",
    pages = "27304--27331",
    ISBN = "979-8-89176-395-1",
}

@book{borg2005modern,
  title     = {Modern Multidimensional Scaling: Theory and Applications},
  author    = {Borg, Ingwer and Groenen, Patrick J. F.},
  series    = {Springer Series in Statistics},
  edition   = {2},
  publisher = {Springer},
  address   = {New York, NY},
  year      = {2005},
  doi       = {10.1007/0-387-28981-X},
  isbn      = {978-0-387-25150-9}
}

@inproceedings{broder1997resemblance,
  title     = {On the Resemblance and Containment of Documents},
  author    = {Broder, Andrei Z.},
  booktitle = {Proceedings. Compression and Complexity of SEQUENCES 1997 (Cat. No.97TB100171)},
  pages     = {21--29},
  year      = {1997},
  publisher = {IEEE Computer Society},
  doi       = {10.1109/SEQUEN.1997.666900}
}

@book{aristotle1995poetics,
  title     = {Poetics},
  author    = {Aristotle},
  editor    = {Halliwell, Stephen},
  series    = {Loeb Classical Library},
  year      = {1995},
  address   = {Cambridge, MA, USA},
  publisher = {Harvard University Press}
}

@incollection{barthes1977introduction,
  title      = {Introduction to the Structural Analysis of Narratives},
  author     = {Barthes, Roland},
  booktitle  = {Image-Music-Text},
  editor     = {Heath, Stephen},
  translator = {Heath, Stephen},
  year       = {1977},
  pages      = {79--124},
  address    = {New York, NY, USA},
  publisher  = {Hill and Wang},
  note       = {Original work published 1966}
}

@incollection{ryan2005tellability,
  title     = {Tellability},
  author    = {Ryan, Marie-Laure},
  booktitle = {Routledge Encyclopedia of Narrative Theory},
  editor    = {Herman, David and Jahn, Manfred and Ryan, Marie-Laure},
  year      = {2005},
  pages     = {589--591},
  address   = {London, UK},
  publisher = {Routledge},
  url = {https://www.routledge.com/Routledge-Encyclopedia-of-Narrative-Theory/Herman-Jahn-Ryan/p/book/9780415775120}
}

@book{labov1972language,
  title     = {Language in the Inner City: Studies in the Black English Vernacular},
  author    = {Labov, William},
  year      = {1972},
  address   = {Philadelphia, PA, USA},
  publisher = {University of Pennsylvania Press},
  url = {https://www.google.co.kr/books/edition/Language_in_the_Inner_City/snEEdFKLJ5cC?hl=ko&gbpv=0}
}

@book{baroni2007tension,
  title     = {La tension narrative: Suspense, curiosit{\'e} et surprise},
  author    = {Baroni, Rapha{\"e}l},
  year      = {2007},
  address   = {Paris, France},
  publisher = {{\'E}ditions du Seuil},
  url       = {https://www.seuil.com/ouvrage/la-tension-narrative-suspense-curiosite-et-surprise-raphael-baroni/9782020906777}
}

@inproceedings{si2025can,
  title={Can {LLM}s {G}enerate {N}ovel {R}esearch {I}deas? {A} {L}arge-{S}cale {H}uman {S}tudy with 100+ {NLP} {R}esearchers},
  author={Si, Chenglei and Yang, Diyi and Hashimoto, Tatsunori},
  booktitle={International Conference on Learning Representations},
  year={2025},
  url = {https://proceedings.iclr.cc/paper_files/paper/2025/hash/ea94957d81b1c1caf87ef5319fa6b467-Abstract-Conference.html}
}

\newpage
\appendix

\renewcommand{\sectionautorefname}{Appendix}
\renewcommand{\subsectionautorefname}{Appendix}
\renewcommand{\subsubsectionautorefname}{Appendix}

\onecolumn

\section{Model Setup}
\label{app:models}
\vspace{-0.5em}

\begin{table}[H]
\vspace{0.05em}
\centering
\begin{threeparttable}
\renewcommand{\arraystretch}{1.1}
\normalsize
\begin{tabularx}{\textwidth}{@{}>{\raggedright\arraybackslash}p{0.22\textwidth}
                            >{\raggedright\arraybackslash}X@{}}
\toprule
\textbf{Model} & \textbf{OpenRouter ID} \\
\midrule
\multicolumn{2}{@{}l}{\textit{Commercial}} \\
Claude Opus 4.7  & \path{anthropic/claude-4.7-opus-20260416} \\
Gemini 3.1 Pro   & \path{google/gemini-3.1-pro-preview-20260219} \\
GPT-5.5          & \path{openai/gpt-5.5-20260423} \\
Grok 4.3         & \path{x-ai/grok-4.3-20260430} \\
Qwen3.6-Plus     & \path{qwen/qwen3.6-plus-04-02} \\
\midrule
\multicolumn{2}{@{}l}{\textit{Open-weight}} \\
DeepSeek-V4-Pro  & \path{deepseek/deepseek-v4-pro-20260423} \\
Kimi K2.6        & \path{moonshotai/kimi-k2.6-20260420} \\
Llama 4 Maverick & \path{meta-llama/llama-4-maverick-17b-128e-instruct} \\
Mistral Large 3  & \path{mistralai/mistral-large-2512} \\
Qwen3.6-35B-A3B  & \path{qwen/qwen3.6-35b-a3b-20260415} \\
\bottomrule
\end{tabularx}
\vspace{-0.3em}
\caption{All models were accessed via OpenRouter\protect\footnotemark; exact snapshot identifiers are provided.
The set spans nine developers and covers frontier general-purpose systems alongside open-weight models of varying scale and training lineage, providing sufficient heterogeneity to probe self-preference across diverse judges and model families.
All models are queried with \texttt{temperature} = 1.0 and \texttt{reasoning\_effort} = \textit{high} where each parameter is supported; unsupported parameters were left at their defaults.}
\label{tab:openrouter_models}
\end{threeparttable}
\vspace{1.0em}
\end{table}
\footnotetext{\url{https://openrouter.ai}}

\section{Evaluation Rubric}
\label{app:rubric}
\vspace{0.05em}

\begin{table}[H]
\centering
\normalsize
\begin{tabular}{@{}>{\centering\arraybackslash}m{2.2cm} p{0.40\linewidth} p{0.40\linewidth}@{}}
\toprule
\multicolumn{1}{c}{\textbf{Dimension}} & \textbf{Anchor at 1 (Low)} & \textbf{Anchor at 7 (High)} \\
\midrule
Originality & The selected constraints lean on familiar story patterns and well-worn tropes, producing a combination that is predictable and conventional. & The selected constraints break from familiar story patterns and well-worn tropes, producing a combination that is unpredictable and fresh. \\[11pt]
Dimensionality & The selected constraints operate independently; altering one does not affect the meaning or function of the others. & The selected constraints are mutually constitutive, each element defining the others' meaning and weight, producing narrative depth. \\[11pt]
Coherence & The selected constraints contain irreconcilable contradictions that no narrative device can resolve within a single coherent story. & The selected constraints integrate seamlessly into a single narrative without logical tensions or unresolved contradictions. \\[11pt]
Tellability & The selected constraints lack a point worth telling---no conflict, question, or stake arises that would make an audience want to hear or tell this story. & The selected constraints carry a point worth telling---a clear conflict, urgent question, or meaningful stake that naturally invites narration and sustains attention. \\
\bottomrule
\end{tabular}
\caption{Four-dimension evaluation rubric with anchor descriptions at the endpoints of a 7-point Likert scale. Coherence asks whether the constraints can share one story, Dimensionality whether they need each other once they do. Theoretical grounding for each dimension is provided in the text below. }
\end{table}

\twocolumn

\paragraph{Theoretical Rationale for the Proposed Rubric.}
Existing narrative evaluation rubrics are built for generated text.
Our task, however, produces selections, not text, so we use a four-dimension rubric that scores a selection directly: \textit{Originality}, \textit{Dimensionality}, \textit{Coherence}, and \textit{Tellability}.
Each of these four aspects of narrative design targets a specific property of the selection instead of asking for a single overall quality score, and the four are meant to measure different things.
We deliberately remove surface properties like style and pacing since they can't be judged before a story is written, and scoring them would bring back the stylistic confound.

We adapt \textit{Originality} and \textit{Coherence} from the Torrance Test of Creative Writing \cite{chakrabarty2024art}, a framework for assessing the creativity of LLM-generated narratives.
While TTCW operates at the surface text level, we modify its construct definitions to selection-level material by rewriting anchors to describe configurations of chosen components rather than executed prose. \textit{Originality} rewards selections that break from familiar patterns and tropes, and \textit{Coherence} rewards selections whose constraints fit together without contradiction.

\textit{Dimensionality} is a selection-level construct we introduce to characterize how chosen narrative components interlock to produce narrative depth — the degree to which each element constitutively shapes the meaning of the others.
We introduce the term, but the idea is old: the classical principle that the parts of a narrative are mutually reinforcing, each taking its meaning from the others \cite{aristotle1995poetics, barthes1977introduction}.
We distinguish it from \textit{Coherence}: \textit{Coherence} asks whether the pieces can share one story, \textit{Dimensionality} whether they need each other once they do. 
%The distinction also shows in the data—the four dimensions behave differently, and Dimensionality shows the largest label effect of any axis (Table 5).

\textit{Tellability} measures whether a selection has a point worth telling—a conflict, question, or stake that answers ``so what?'' The idea originates from Labov's study of natural storytelling \cite{labov1972language}, where an event is narratable only if it earns the telling. It later came to be treated as a core category in narrative theory \cite{ryan2005tellability, baroni2007tension}. It is independent of the other three since a selection can be original, coherent, and dimensional and still give a reader no reason to care, making the story lack \textit{Tellability}.

LLM judges are usually run in one of two ways: pairwise comparison or direct scoring \cite{pombal2026self}. For creative work, ``which one is better'' is a less meaningful task because two very different selections can both be defensible. Thus we score each selection on its own, on a 7-point scale with described endpoints. Scoring selections individually also gives us an absolute number per selection, which is what lets the Experiment 2 label manipulation move a single rating instead of only a ranking.

We did not validate the rubric against human annotation. It functions as a held-constant comparative frame, since our claims concern differences in how a fixed instrument is applied across conditions, not absolute quality scores. Validating it against human judgments would be needed before using it to score quality in absolute terms.

\onecolumn

\section{Evaluation Run Template}\label{app:experiment_1}
\vspace{-0.0em}
\subsection{Experiment 1: Blind Evaluation}
\vspace{-0.5em}
\begin{figure}[H]
\centering
\footnotesize

\begin{tcolorbox}[
  colback=gray!6, colframe=gray!30, coltitle=black,
  title=\textbf{Run metadata}, fonttitle=\footnotesize\bfseries,
  boxrule=0.4pt, arc=2pt,
  left=6pt, right=6pt, top=4pt, bottom=4pt
]
\begin{tabular}{@{}ll@{}}
Judge                & \texttt{Kimi K2.6} \\
Selector             & \texttt{GPT-5.5} \\
\texttt{is\_self}    & \texttt{False} \\
Repetition           & 1 of 3 \\
Dimension order      & Originality $\to$ Coherence $\to$ Tellability $\to$ Dimensionality \\
\end{tabular}
\end{tcolorbox}

\vspace{4pt}

% ===== Input prompt =====
\begin{tcolorbox}[
  colback=cyan!4, colframe=cyan!20, coltitle=black,
  title=\textbf{Input prompt}, fonttitle=\footnotesize\bfseries,
  boxrule=0.4pt, arc=2pt, breakable,
  left=6pt, right=6pt, top=4pt, bottom=4pt
]
You will rate a selection of 20 narrative constraints (chosen from a larger pool) on four dimensions, each on a 1--7 Likert scale.

\smallskip
\textit{\textcolor{gray}{[Rubric block: four dimensions, each anchored at 1 and 7; see Table 9 in~\autoref{app:rubric}. Dimensions are presented in the randomized order recorded in the metadata above.]}}
\smallskip

\begin{itemize}[leftmargin=*, itemsep=0pt, topsep=0pt]
\item Use the full 1--7 range.
\item Respond in English.
\item For each dimension, provide a score (1--7) and 2--3 sentences explaining the reason for that score.
\end{itemize}

\smallskip
\textbf{Selection:}\\
\textit{\textcolor{gray}{[20 selected constraints listed in randomized order]}}

\smallskip
Respond with JSON only:
\begin{quote}\ttfamily\footnotesize
\{"originality": \{"score": <int>, "reasoning": "<2-3 sentences>"\}, "coherence": \{\ldots\}, "tellability": \{\ldots\}, "dimensionality": \{\ldots\}\}
\end{quote}
\end{tcolorbox}

\vspace{4pt}

% ===== Model output =====
\begin{tcolorbox}[
  colback=violet!4, colframe=violet!15, coltitle=black,
  title=\textbf{Model output}, fonttitle=\footnotesize\bfseries,
  boxrule=0.4pt, arc=2pt, breakable,
  left=6pt, right=6pt, top=4pt, bottom=4pt
]
\ttfamily
\{\\
\hspace*{1em}"originality":~~~~~\{"score": 5, "reasoning": "\textit{\rmfamily<\ldots>}"\},\\
\hspace*{1em}"coherence":~~~~~~~\{"score": 4, "reasoning": "\textit{\rmfamily<\ldots>}"\},\\
\hspace*{1em}"tellability":~~~~~\{"score": 6, "reasoning": "\textit{\rmfamily<\ldots>}"\},\\
\hspace*{1em}"dimensionality":~~\{"score": 3, "reasoning": "\textit{\rmfamily<\ldots>}"\}\\
\}
\end{tcolorbox}

\caption{Schematic of an Experiment~1 evaluation run (Kimi K2.6 evaluating a GPT-5.5 selection). The model output reports actual scores from this run; per-dimension reasoning text is omitted.}
\label{fig:eval_template}
\end{figure}

\subsection{Experiment 2: Labeled Evaluation}
\vspace{-0.5em}

\begin{figure}[H]
\centering
\footnotesize
\begin{tcolorbox}[
  colback=gray!6, colframe=gray!30, coltitle=black,
  title=\textbf{Run metadata}, fonttitle=\footnotesize\bfseries,
  boxrule=0.4pt, arc=2pt,
  left=6pt, right=6pt, top=4pt, bottom=4pt
]
\begin{tabular}{@{}ll@{}}
Judge                & \texttt{Gemini 3.1 Pro} \\
Selector             & \texttt{Qwen3.6-Plus} \\
\texttt{is\_self}    & \texttt{False} \\
Condition            & \texttt{FL-self} \\
Repetition           & 3 of 3 \\
Dimension order      & Originality $\to$ Dimensionality $\to$ Coherence $\to$ Tellability \\
\end{tabular}
\end{tcolorbox}
\vspace{4pt}
\begin{tcolorbox}[
  colback=cyan!4, colframe=cyan!20, coltitle=black,
  title=\textbf{Input prompt}, fonttitle=\footnotesize\bfseries,
  boxrule=0.4pt, arc=2pt, breakable,
  left=6pt, right=6pt, top=4pt, bottom=4pt
]
The following selection of 20 narrative constraints is \textbf{your own selection} from a larger pool of 200, made earlier as most useful for writing a single fictional narrative.
\smallskip

\textit{\textcolor{gray}{[Rubric block, instructions, selection list, and JSON format identical to Experiment~1; see Figure~\ref{fig:eval_template}.]}}
\end{tcolorbox}
\caption{Schematic of an Experiment~2 evaluation run (Gemini 3.1 Pro evaluating a Qwen3.6-Plus selection under the FL-self condition: the selection is labeled as the judge's own, though it was produced by another model).  The input prompt differs from Experiment~1 only in the opening label phrase (bold); all other components are identical.}
\label{fig:eval_template_exp2}
\end{figure}

\clearpage
\twocolumn[{
\begin{minipage}{\textwidth}

\section{k-NN Source-Model Classification}
\label{sec:appendix_knn_perm}

\begin{center}
\includegraphics[
    width=0.92\textwidth,
    trim={0 0 0 0.7cm},
    clip
]{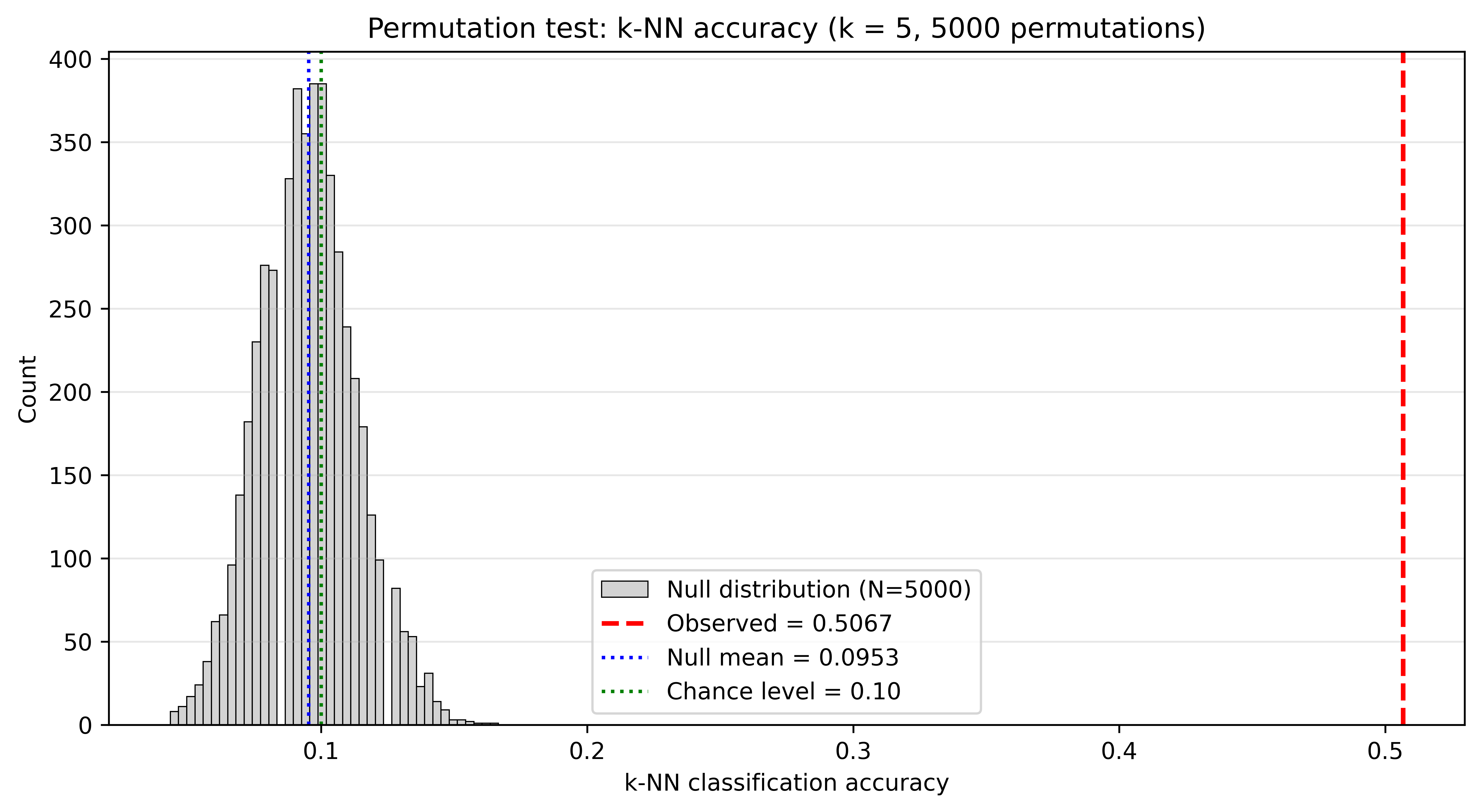}
\end{center}

\vspace{-0.5em}

\refstepcounter{figure}
\label{fig:knn_perm}
{\small
\noindent Figure~\thefigure:
Label-permutation null distribution for the k-NN source-model classifier ($k=5$).
The null is constructed by randomly shuffling model labels 5{,}000 times and recomputing leave-one-out k-NN accuracy at each iteration.
The observed accuracy of 0.507 falls far above the null distribution (null mean $= 0.095$, close to the 0.10 chance baseline; $p < .001$),
confirming that selection-based source identification substantially exceeds chance.
\par}

\vspace{1.0em}

\end{minipage}
}]

\vspace*{-1.0em}

% Table A1: k-NN accuracy summary across k values
\begin{table}[H]
\centering
\renewcommand{\arraystretch}{1.1}
\small
\setlength{\tabcolsep}{5pt}
\begin{tabular}{@{}cccccc@{}}
\toprule
$k$ & \textbf{Accuracy} & \textbf{Correct/$N$} & \textbf{Macro-F1} & $\kappa$ & \textbf{$p$} \\
\midrule
1 & 0.477 & 143/300 & 0.427 & 0.419 & *** \\
3 & 0.467 & 140/300 & 0.405 & 0.407 & *** \\
5 & \textbf{0.507} & \textbf{152/300} & \textbf{0.448} & \textbf{0.452} & *** \\
7 & 0.493 & 148/300 & 0.424 & 0.437 & *** \\
\bottomrule
\end{tabular}
\caption{Leave-one-out $k$-NN classification of selection profiles by source model ($N=300$, chance $=0.10$, 10 classes).
Cohen's $\kappa$ ranges from 0.41 to 0.45, indicating \emph{moderate} chance-corrected agreement.
$k=5$ is used as the primary configuration.
$^{***} p < .001$ (binomial test against $H_0: p = 1/10$).}
\label{tab:knn_accuracy}
\end{table}

\vspace*{-1.0em}

% =====================================================================
\begin{table}[H]
\centering
\renewcommand{\arraystretch}{1.1}
\small
\setlength{\tabcolsep}{4pt}
\begin{tabular}{@{}lcccrrr@{}}
\toprule
\textbf{Model} & \textbf{TP} & \textbf{FP} & \textbf{FN} & \textbf{Prec.} & \textbf{Rec.} & \textbf{F1} \\
\midrule
Claude Opus 4.7   & 25 & 10 &  5 & 0.71 & 0.83 & \textbf{0.77} \\
Llama 4 Maverick  & 30 & 22 &  0 & 0.58 & 1.00 & 0.73 \\
GPT-5.5           & 30 & 37 &  0 & 0.45 & 1.00 & 0.62 \\
Kimi K2.6         & 20 & 18 & 10 & 0.53 & 0.67 & 0.59 \\
Qwen3.6-Plus      & 20 & 24 & 10 & 0.45 & 0.67 & 0.54 \\
DeepSeek-V4-Pro   &  7 &  6 & 23 & 0.54 & 0.23 & 0.33 \\
Gemini 3.1 Pro    &  8 & 12 & 22 & 0.40 & 0.27 & 0.32 \\
Qwen3.6-35B-A3B   &  5 & 10 & 25 & 0.33 & 0.17 & 0.22 \\
Grok 4.3          &  4 &  9 & 26 & 0.31 & 0.13 & 0.19 \\
Mistral Large 3   &  3 &  0 & 27 & 1.00 & 0.10 & 0.18 \\
\bottomrule
\end{tabular}
\caption{Per-model $k$-NN classification report ($k=5$, $n=30$ per model), sorted by F1 descending.
Two asymmetric patterns stand out: Llama 4 Maverick and GPT-5.5 achieve perfect recall but lower precision, whereas Mistral Large 3 achieves perfect precision but is rarely selected as the prediction.}
\label{tab:knn_per_model}
\end{table}

\vspace*{3.5em}

\section{Within-Model Jaccard Similarity}
\label{app:intra_jaccard}

\begin{table}[H]
\renewcommand{\arraystretch}{1.1}
\centering
\small
\setlength{\tabcolsep}{4pt}
\begin{tabular}{lcccc c}
\toprule
\textbf{Model}
  & \textbf{Mean}
  & \textbf{SD}
  & \textbf{Min}
  & \textbf{Max}
  & \\
\midrule
\multicolumn{6}{l}{\textit{Commercial}} \\
Claude Opus 4.7  & 0.218 & 0.086 & 0.026 & 0.482 \\
Gemini 3.1 Pro   & 0.190 & 0.093 & 0.026 & 0.600 \\
GPT-5.5          & 0.405 & 0.092 & 0.143 & 0.667 \\
Grok 4.3         & 0.120 & 0.060 & 0.000 & 0.333 \\
Qwen3.6-Plus     & 0.235 & 0.079 & 0.026 & 0.539  \\
\midrule
\multicolumn{6}{l}{\textit{Open-weight}} \\
DeepSeek-V4-Pro  & 0.136 & 0.062 & 0.000 & 0.429 \\
Kimi K2.6        & 0.221 & 0.069 & 0.081 & 0.429 \\
Llama 4 Maverick & 0.247 & 0.091 & 0.053 & 0.600 \\
Mistral Large 3  & 0.072 & 0.040 & 0.000 & 0.212 \\
Qwen3.6-35B-A3B  & 0.153 & 0.064 & 0.000 & 0.379 \\
\midrule
\textit{Random null}
  & \textit{0.054}
  & \multicolumn{3}{c}{---}
  & \\
\bottomrule
\end{tabular}
\caption{Within-model selection consistency measured by pairwise Jaccard similarity, computed over all $\binom{30}{2} = 435$ unique pairs from 30 selection runs per model.
The random null is the expected mean Jaccard under uniform random selection from 200 constraints
(10{,}000 permutations; mean $= 0.054$, 95\%~CI: $[0.051, 0.058]$).
All ten models exceed the random null at $p <
.0001$ (one-tailed permutation test).}
\label{tab:intra_jaccard}
\end{table}

\clearpage
\onecolumn

\section{Self–Other Comparison by Judge Across Rubric Dimensions}
\label{app:raw_dims}

\begin{table}[H]
\centering
\small
\setlength{\tabcolsep}{5pt}
\renewcommand{\arraystretch}{1.15}
\begin{tabular}{@{}l rrr rrr rrr rrr@{}}
\toprule
& \multicolumn{3}{c}{\textbf{Originality}}
& \multicolumn{3}{c}{\textbf{Dimensionality}}
& \multicolumn{3}{c}{\textbf{Coherence}}
& \multicolumn{3}{c}{\textbf{Tellability}} \\
\cmidrule(lr){2-4}\cmidrule(lr){5-7}\cmidrule(lr){8-10}\cmidrule(lr){11-13}
\textbf{Judge}
  & \textbf{Self} & \textbf{Other} & \textbf{$\Delta$}
  & \textbf{Self} & \textbf{Other} & \textbf{$\Delta$}
  & \textbf{Self} & \textbf{Other} & \textbf{$\Delta$}
  & \textbf{Self} & \textbf{Other} & \textbf{$\Delta$} \\
\midrule
Kimi K2.6        & 3.43 & 4.47 & $-1.04$ & 5.44 & 3.40 & $+2.05$ & 4.72 & 2.44 & $+2.28$ & 5.93 & 4.85 & $+1.08$ \\
GPT-5.5          & 4.46 & 4.74 & $-0.28$ & 5.99 & 5.16 & $+0.83$ & 5.83 & 3.89 & $+1.94$ & 6.82 & 5.87 & $+0.95$ \\
DeepSeek-V4-Pro  & 5.68 & 5.07 & $+0.60$ & 4.40 & 4.02 & $+0.38$ & 4.02 & 3.48 & $+0.54$ & 5.33 & 4.78 & $+0.56$ \\
Claude Opus 4.7  & 3.84 & 3.87 & $-0.03$ & 4.13 & 3.23 & $+0.91$ & 2.69 & 2.13 & $+0.56$ & 5.36 & 4.77 & $+0.58$ \\
Qwen3.6-Plus     & 4.74 & 5.00 & $-0.26$ & 4.96 & 4.48 & $+0.47$ & 3.53 & 3.16 & $+0.37$ & 5.69 & 5.30 & $+0.39$ \\
Gemini 3.1 Pro   & 3.76 & 4.63 & $-0.88$ & 4.94 & 4.34 & $+0.60$ & 2.86 & 2.27 & $+0.59$ & 6.10 & 5.59 & $+0.51$ \\
Llama 4 Maverick & 5.39 & 4.89 & $+0.50$ & 5.89 & 5.83 & $+0.06$ & 4.90 & 5.37 & $-0.47$ & 5.86 & 5.78 & $+0.07$ \\
Mistral Large 3  & 5.98 & 5.70 & $+0.28$ & 6.26 & 6.38 & $-0.12$ & 5.00 & 5.41 & $-0.41$ & 6.52 & 6.60 & $-0.07$ \\
Grok 4.3         & 4.97 & 4.42 & $+0.55$ & 3.02 & 4.07 & $-1.04$ & 2.06 & 3.51 & $-1.45$ & 4.69 & 5.41 & $-0.72$ \\
Qwen3.6-35B-A3B  & 3.73 & 4.62 & $-0.89$ & 3.96 & 4.93 & $-0.97$ & 2.29 & 3.47 & $-1.18$ & 5.02 & 5.59 & $-0.56$ \\
\bottomrule
\end{tabular}
\caption{Raw self--other comparison per judge across all four rubric dimensions. $\Delta = \text{Self} - \text{Other}$. Confounds are not controlled.}
\end{table}

\section{Average Score by Judge and Selector}
\label{fig:judge_selector_heatmap}
\vspace{-0.0em}

\begin{figure}[H]
\centering
\includegraphics[width=\textwidth,height=0.525\textheight,keepaspectratio]{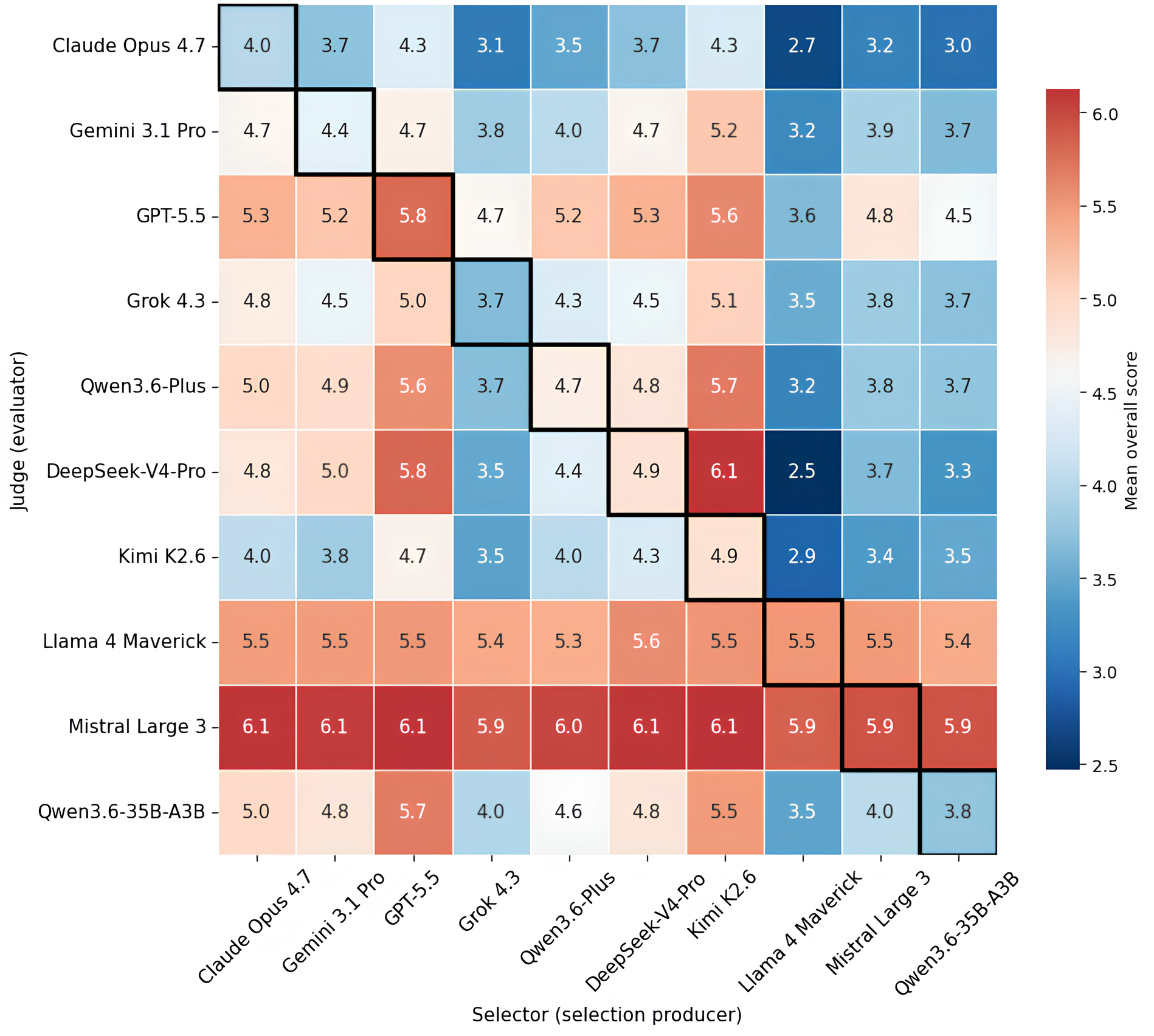}
\caption{Average score assigned by each judge (rows) to each selector's selections (columns).
Diagonal cells indicate self-evaluation.
Row-level variation reflects judge-level severity differences; column-level variation reflects selection quality.
Selectors that receive consistently high scores across judges---notably GPT-5.5 and Kimi K2.6---also score highly on the diagonal, suggesting that the raw self--other gap is driven in part by selection quality rather than genuine self-preference.}
\end{figure}

\twocolumn
\section{Random-Slope Robustness Check}
\label{app:random_slope}

The random intercepts of the model in \S\ref{sec:method_exp1_analysis} absorb judge severity, selection quality, and the non-independence of repeated ratings.
However, they cannot capture whether a judge discriminates among selections at all. 
Under the single $\beta_S$ shared by all judges, the two non-discriminating judges---whose scores carry no self--other signal---still enter the estimate, as if their near-zero gaps were evidence about self-preference.
To model judge heterogeneity directly, we refit the model on the full 10-judge sample with a random slope for \textit{is\_self} by judge:

\begin{equation*}
\small
\begin{split}
\texttt{score} = \beta_0 &+ \beta_S \cdot \texttt{is\_self} + (1 \,|\, \texttt{judge}) \\
&+ (0 + \texttt{is\_self} \,|\, \texttt{judge}) \\
&+ (1 \,|\, \texttt{selector}) + (1 \,|\, \texttt{selection\_id})
\end{split}
\end{equation*}

\begin{table}[h]
\centering
\small
\setlength{\tabcolsep}{5pt}
\renewcommand{\arraystretch}{1.1}
\begin{tabular}{@{}l r c c c r@{}}
\toprule
& & \textbf{M0} & \multicolumn{2}{c}{\textbf{M1}} & \\
\cmidrule(lr){3-3} \cmidrule(lr){4-5}
\textbf{Dim.} & \multicolumn{1}{c}{$\beta_S$} & $p$ & 95\% CI & $p$ & \multicolumn{1}{c}{$\tau$} \\
\midrule
Avg. & $+.181$ & $<.001$ & $[-.08, .44]$ & $.173$ & $.41$ \\
Org. & $-.144$ & $<.001$ & $[-.41, .12]$ & $.284$ & $.41$ \\
Dim. & $+.316$ & $<.001$ & $[-.08, .71]$ & $.120$ & $.63$ \\
Coh. & $+.276$ & $<.001$ & $[-.05, .60]$ & $.098$ & $.52$ \\
Tel. & $+.278$ & $<.001$ & $[-.07, .63]$ & $.116$ & $.55$ \\
\bottomrule
\end{tabular}
\caption{Self-preference coefficient $\beta_S$ on the full 10-judge sample under the intercept-only model (M0) and the random-slope model (M1). $\tau$ = between-judge SD of the self effect under M1. Point estimates are shared across the two specifications; M0 confidence intervals appear in the ``Full (10 judges)'' column of Table~\ref{tab:confound}.}
\label{tab:random_slope}
\end{table}
Point estimates are unchanged, but the standard errors now account for between-judge variability: no dimension shows a significant average self-preference, including \textit{Originality}. The heterogeneity is substantial relative to the point estimates ($\tau = 0.41$--$0.63$), indicating that the self effect varies markedly across judges rather than reflecting a shared bias.

\vspace{13em}
\section{Experiment 2 Estimates Excluding Low-Discrimination Judges}
\label{app:exp2_mixed_excl}

\begin{table}[h]
\centering
\renewcommand{\arraystretch}{1.1}
\small
\setlength{\tabcolsep}{4pt}
\begin{tabular}{@{}ccccc@{}}
\toprule
\textbf{Dim.} & \textbf{Effect} & \textbf{Estimate} & \textbf{95\% CI} & $\mathbf{p}$ \\
\midrule
              & $\beta_L$    & $\mathbf{+0.47^{***}}$ & $[+0.43, +0.52]$ & $<.001$ \\
\textbf{Avg.} & $\beta_A$    & $+0.08$                & $[-0.07, +0.22]$ & $.293$  \\
              & $\beta_{LA}$ & $-0.02$                & $[-0.09, +0.05]$ & $.638$  \\
\midrule
     & $\beta_L$    & $\mathbf{+0.57^{***}}$ & $[+0.49, +0.65]$ & $<.001$ \\
Org. & $\beta_A$    & $-0.12$                & $[-0.27, +0.02]$ & $.089$  \\
     & $\beta_{LA}$ & $-0.03$                & $[-0.14, +0.08]$ & $.595$  \\
\midrule
     & $\beta_L$    & $\mathbf{+0.68^{***}}$ & $[+0.59, +0.76]$ & $<.001$ \\
Dim. & $\beta_A$    & $+0.15$                & $[-0.06, +0.35]$ & $.164$  \\
     & $\beta_{LA}$ & $-0.04$                & $[-0.17, +0.08]$ & $.468$  \\
\midrule
     & $\beta_L$    & $\mathbf{+0.27^{***}}$ & $[+0.20, +0.33]$ & $<.001$ \\
Coh. & $\beta_A$    & $+0.15$                & $[-0.09, +0.39]$ & $.214$  \\
     & $\beta_{LA}$ & $-0.03$                & $[-0.12, +0.06]$ & $.538$  \\
\midrule
     & $\beta_L$    & $\mathbf{+0.38^{***}}$ & $[+0.31, +0.45]$ & $<.001$ \\
Tel. & $\beta_A$    & $+0.13$                & $[-0.03, +0.28]$ & $.101$  \\
     & $\beta_{LA}$ & $+0.04$                & $[-0.06, +0.13]$ & $.462$  \\
\bottomrule
\end{tabular}
\caption{Robustness check: fixed-effect estimates from the mixed-effects models for Experiment~2 after excluding the two low-discrimination judges (Llama 4 Maverick and Mistral Large 3). Model specification identical to Table~\ref{tab:exp2_mixed}.
$\beta_L$ captures the displayed-label effect; $\beta_A$ the actual-source effect (expected $\approx 0$ under successful quality matching); $\beta_{LA}$ their interaction. $^{***} p < .001$. 
The label effect ($\beta_L$) remains the only significant fixed effect on every dimension, and its magnitude is comparable to or slightly larger than the full-sample estimate, confirming that the main findings are not driven by the two excluded judges.}
\end{table}

\end{document}